\documentclass{article} %
\usepackage{iclr2027_conference,times}

\usepackage{amsmath,amsfonts,bm}

\def\eqref#1{equation~\ref{#1}}

\def\1{\bm{1}}

\DeclareMathAlphabet{\mathsfit}{\encodingdefault}{\sfdefault}{m}{sl}
\SetMathAlphabet{\mathsfit}{bold}{\encodingdefault}{\sfdefault}{bx}{n}

\usepackage{amsthm}
\newtheorem{proposition}{Proposition}

\usepackage[normalem]{ulem}
\usepackage{graphicx}
\usepackage{booktabs}
\usepackage{multirow}
\usepackage[table]{xcolor}
\usepackage{tikz}
\usetikzlibrary{patterns.meta}
\usepackage{hyperref}
\usepackage{url}
\usepackage{algorithm}
\usepackage{algpseudocode}
\usepackage{subcaption}
\usepackage{placeins}
\algblockdefx[On]{On}{EndOn}[1]{\textbf{on} #1 \textbf{do}}{\textbf{end on}}

\usetikzlibrary{tikzmark,fit}
\definecolor{AlgAllocation}{HTML}{B66A20}
\definecolor{AlgRefinement}{HTML}{276A92}
\definecolor{AlgContext}{HTML}{76558C}
\newcommand{\algblocktitle}[3][0pt]{%
  \expandafter\gdef\csname algblocklabel#2\endcsname{#3}%
  \Statex\hspace{#1}\tikzmarknode{#2-top}{%
    \makebox[\dimexpr\linewidth-#1\relax][l]{\strut}}}
\newcommand{\algblockend}[1]{\tikzmarknode{#1-bottom}{\strut}}
\newcommand{\algblockbox}[2]{%
  \begin{tikzpicture}[remember picture,overlay]
    \node[fit=(#1-top.south west)(#1-top.south east)(#1-bottom),
      draw=#2,fill=#2,fill opacity=0.05,draw opacity=1,
      rounded corners=3pt,line width=0.6pt,inner xsep=4pt,inner ysep=2pt]
      (#1-box) {};
    \node[anchor=east,fill=white,text=#2,inner xsep=4pt,inner ysep=1pt,
      font=\scriptsize\sffamily\bfseries]
      at ([xshift=-7pt]#1-box.north east) {\csname algblocklabel#1\endcsname};
  \end{tikzpicture}}

\title{FOCUS: Active Critic Learning for Long-Horizon Language Model Agents}

\author{%
  Zhengping Jiang\qquad
  William Jurayj\qquad
  Liaoyaqi Wang\qquad
  Chunsheng Zuo\\
  \textbf{Anqi Liu\qquad
  Benjamin Van Durme}\\
  Johns Hopkins University\\
  \texttt{\{zjiang31, aliu.cs, vandurme\}@jhu.edu}%
}

\definecolor{OmitFill}{gray}{0.94}
\newcommand{\omitbench}{\multicolumn{3}{c}{\cellcolor{OmitFill}}}
\newcommand{\omitbenchlabel}[1]{%
  \multicolumn{3}{c}{\cellcolor{OmitFill}%
    \normalfont\itshape\color{black!55}#1}}

\definecolor{ToolFill}{gray}{0.935}
\definecolor{ToolInk}{gray}{0.22}
\newcommand{\tool}[1]{%
  \tikz[baseline=(t.base)]\node[fill=ToolFill, rounded corners=1.6pt,
    inner xsep=2.2pt, inner ysep=1.1pt, outer sep=0pt](t)
    {\ttfamily\small\color{ToolInk}\vphantom{Agy}#1};}

\usepackage{array}
\usepackage[most]{tcolorbox}
\definecolor{PromptFill}{gray}{0.985}
\definecolor{PromptRule}{gray}{0.80}
\definecolor{PromptLabel}{gray}{0.45}
\definecolor{FreeFormAccent}{HTML}{475569}
\definecolor{StructAccent}{HTML}{0F766E}
\newtcolorbox{promptbox}[3][FreeFormAccent]{%
  enhanced, breakable, sharp corners,
  colback=PromptFill, colframe=PromptRule, boxrule=0.4pt,
  borderline west={2.2pt}{0pt}{#1},
  colbacktitle=#1!9!white, coltitle=#1!80!black,
  fonttitle=\sffamily\bfseries\footnotesize,
  title={#2\hfill{\mdseries\ttfamily\scriptsize #3}},
  toptitle=1.5pt, bottomtitle=1.5pt, titlerule=0.4pt,
  left=7pt, right=7pt, top=5pt, bottom=5pt,
  before skip=6pt, after skip=9pt,
  before upper={\footnotesize\ttfamily\raggedright\frenchspacing
    \setlength{\parskip}{0.5em}\setlength{\parindent}{0pt}}}
\newcommand{\promptpart}[1]{%
  \par\vspace{0.15em}%
  {\normalfont\sffamily\scriptsize\color{PromptLabel}\MakeUppercase{#1}}%
  \par\nobreak\vspace{-0.3em}}
\newcommand{\pslot}[1]{%
  \tikz[baseline=(s.base)]\node[fill=black!8, rounded corners=1.2pt,
    inner xsep=2.2pt, inner ysep=0.9pt, outer sep=0pt](s)
    {\normalfont\sffamily\itshape\scriptsize\color{black!62}\vphantom{Agy}#1};}
\newcommand{\phead}[1]{{\bfseries\#\# #1}}
\newcommand{\us}{{\ttfamily\char95}}

\definecolor{TraceAccent}{HTML}{4C72B0}   %
\definecolor{TraceKeyFill}{HTML}{FCEBB6}
\definecolor{TraceGapInk}{HTML}{B42318}
\definecolor{TracePartialInk}{HTML}{B45309}
\newtcolorbox{tracebox}[2]{%
  enhanced, sharp corners,  %
  colback=PromptFill, colframe=PromptRule, boxrule=0.4pt,
  borderline west={2.2pt}{0pt}{TraceAccent},
  colbacktitle=TraceAccent!9!white, coltitle=TraceAccent!80!black,
  fonttitle=\sffamily\bfseries\footnotesize,
  title={#1\hfill{\mdseries\ttfamily\scriptsize #2}},
  toptitle=1.5pt, bottomtitle=1.5pt, titlerule=0.4pt,
  left=7pt, right=7pt, top=4pt, bottom=5pt,
  before skip=6pt, after skip=9pt,
  before upper={\footnotesize\raggedright\frenchspacing
    \setlength{\parindent}{0pt}\setlength{\parskip}{0pt}}}
\newcommand{\tracesep}{\enspace{\color{PromptRule}\textbar}\enspace}
\newcommand{\tracemeta}[4]{%
  {\sffamily\scriptsize\color{black!72}#1\tracesep #2\tracesep #3\tracesep #4}\par}
\newcommand{\tracepart}[1]{%
  \par\vspace{0.5em}%
  {\sffamily\scriptsize\color{PromptLabel}\MakeUppercase{#1}}%
  \par\nobreak\vspace{0.15em}}
\newcommand{\tracesteplabel}[1]{%
  \makebox[2.1em][l]{\sffamily\scriptsize\color{PromptLabel}#1}}
\newcommand{\tracestep}[3]{%
  \par\tracesteplabel{#1}#2\par\nobreak
  {\leftskip=2.1em\ttfamily\scriptsize\color{black!80}#3\par}\vspace{0.3em}}
\newcommand{\tracequote}[1]{%
  {\leftskip=2.1em\ttfamily\scriptsize\color{black!80}#1\par}}
\newcommand{\tracecall}[2]{%
  \par\hangindent=2.1em\hangafter=1
  \tracesteplabel{#1}#2\par
  \vspace{0.15em}}
\newcommand{\tracecallsep}{\enspace}
\newcommand{\tracefold}[1]{%
  {\sffamily\scriptsize\bfseries\color{TraceAccent!80!black}compaction}%
  \enspace{\sffamily\scriptsize\color{PromptLabel}#1}}
\newcommand{\tracesubmit}[3]{%
  \par{\leftskip=2.1em\rightskip=0pt plus 0pt\parfillskip=0pt
  \tool{submit}\,{\ttfamily\scriptsize #1}%
  \unskip\nobreak\hfil\penalty50\hskip1.5em\null\nobreak\hfill
  {\sffamily\scriptsize\color{PromptLabel}#2}\enspace{\ttfamily\scriptsize #3}\par}}
\newcommand{\tracebad}[1]{\textcolor{TraceGapInk}{\bfseries #1}}
\newcommand{\tracepartial}[1]{\textcolor{TracePartialInk}{\bfseries #1}}
\newcommand{\tracekey}{\bgroup
  \markoverwith{\textcolor{TraceKeyFill}{\rule[-0.6ex]{2pt}{2.3ex}}}\ULon}
\newcommand{\tracegapul}{\bgroup
  \markoverwith{\textcolor{TraceGapInk}{\rule[-0.55ex]{2pt}{0.5pt}}}\ULon}
\newcommand{\tracegap}[1]{{\color{TraceGapInk}\tracegapul{#1}}}

\iclrfinalcopy %
\begin{document}

\maketitle
\raggedbottom
\lhead{} %

\begin{abstract}
Sparse outcome rewards make it difficult for long-horizon language-model
agents to assign credit to earlier actions. Learned critics provide
value estimates for intermediate states, but these estimates can be inaccurate, while
estimating intermediate values with repetitive continuation samples from points
within a trajectory incurs substantial costs. To mitigate these issues,
we introduce FOCUS, an RL method that uses selective branching to help the
critic network learn its value estimator. The rollout controller facilitates
active learning for the critic by sampling continuations from states that are expected to most reduce the critic's
estimation error. These refined value estimates then score the segment
between consecutive states with a TD($\lambda$) advantage, while the
sampled evidence supplies training targets for the critic.
FOCUS supports long-horizon agentic reinforcement learning through a modular context-management interface that decouples context management from rollout allocation.
Our experiments show that using compaction-induced segment boundaries for value estimation and credit assignment,
FOCUS outperforms GRPO and VinePPO with matched per-step
generation budgets on legal reasoning, deep search, and code localization benchmarks, improving held-out accuracy at Qwen3.5-9B scale by
7.2 points over the strongest baseline on BrowseComp-Plus (55.4 vs.\ 48.2)
and by 4.1 points on DeonticBench (56.5 vs.\ 52.4).
\end{abstract}

\section{Introduction}
\label{sec:intro}

Language-model agents are increasingly trained to solve tasks that unfold
over dozens or hundreds of tool calls: resolving an issue in a real
repository~\citep{jimenez2024swebench,cursor2026composer2}, completing a
task on a live website~\citep{zhou2024webarena}, or answering a question
that requires many rounds of search and
reading~\citep{chen-etal-2026-browsecomp,kimi2026k3}. Reinforcement
Learning with Verified Rewards (RLVR) is the standard way to
improve such agents, but long trajectories mean both sampling and credit
assignment are costly. A single verifiable reward must be distributed
across every token within a rollout even when many had little impact on
the final outcome. To learn efficiently, we need proper intermediate
value estimates under a limited rollout budget.

These estimates can come from a learned critic network~\citep{konda1999actor}
or through Monte Carlo approximation from a given state~\citep{shao2024deepseekmath}.
A learned critic reuses information across trajectories, but its predictions
can be inaccurate. Tree-structured rollouts branch at intermediate states to
refine value estimates~\citep{kazemnejad2024vineppo,hou2025treerl,li2025treepo,guo2025spo,ji2025treegrpo}.
However, sampling additional language model continuations for each estimate incurs substantial
costs, raising the question of how to allocate a rollout budget to improve
value estimation most efficiently.

We treat refining value estimates via additional continuation branching as
an active learning problem for a critic network: continuations should be
sampled selectively from states where additional evidence is expected to
improve value estimation the most. We introduce \textbf{FOCUS:}
\textbf{F}ocused \textbf{O}ptimization on \textbf{C}ritical
\textbf{U}nderperforming \textbf{S}egments, a rollout controller that runs
each episode as a search tree. FOCUS combines the critic's predictions
with evidence from sampled continuations and allocates further rollouts
according to their estimated marginal reduction in value-estimation
error~\citep{hay2012selecting,kaufmann2017baimcts}. These refined values
provide stepwise credit for the policy, while the acquired evidence
supplies training targets for the critic.

\begin{figure}[t]
  \centering
  \includegraphics[width=\linewidth]{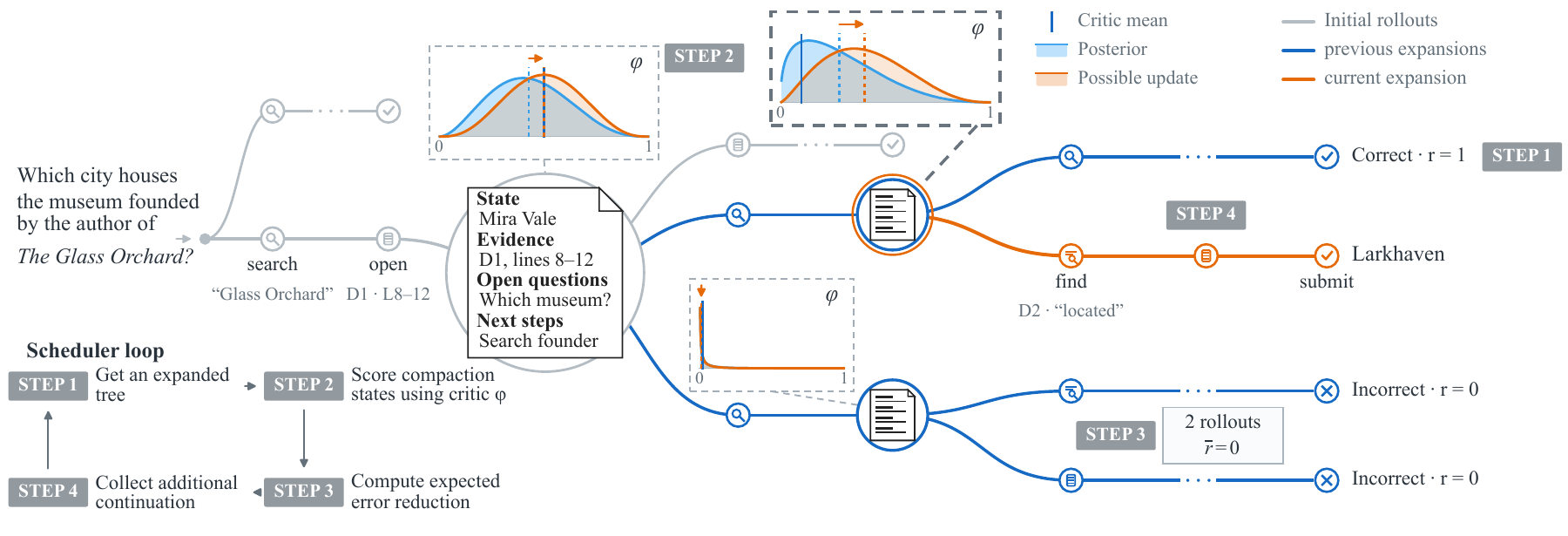}
  \caption{Overview of the proposed FOCUS rollout scheduler for a deep-research-style problem.
  FOCUS continues working on a rollout tree. At each iteration, FOCUS \textbf{STEP 1:} takes an expanded tree; \textbf{STEP 2:} scores each compaction-induced state with the critic; \textbf{STEP 3:} compares against outcomes to decide critic quality (\autoref{sec:method-credit}); \textbf{STEP 4:} sources additional continuations that are expected to correct critic errors most efficiently (\autoref{sec:method-search}). With this active learning loop for rollout acquisition, FOCUS provides a more accurate value estimate with limited rollout budgets (\autoref{sec:value-refinement}).
  }
  \label{fig:pipeline}
\end{figure}

In long-horizon tasks, context compaction~\citep{cursor2026selfsummarization,harvey2026tenet,sun2025contextfolding,li2026compactionrl},
where the agent resumes from a compacted account of its previous actions
and observations, provides a natural structure for applying selective
rollouts: each segment between compactions becomes a \textit{step}, at which step-level rewards
and values can be calculated (Figure~\ref{fig:pipeline}). However, FOCUS does not
depend on a particular context-management protocol; its modular design
separates rollout allocation, context management, and credit
assignment.\footnote{Code: \url{https://github.com/zipJiang/FOCUS}.}
Appendix~\ref{app:execution} details the implementation, and
Section~\ref{sec:related} discusses related work.

We evaluate FOCUS on DeonticBench~\citep{dou2026deonticbench},
BrowseComp-Plus~\citep{chen-etal-2026-browsecomp}, and
LocBench~\citep{chen2025locagent}. FOCUS outperforms GRPO and VinePPO on all
three benchmarks under a matched per-step generation budget. With
Qwen3.5-9B, FOCUS improves held-out accuracy over the strongest baseline
by 7.2 points on BrowseComp-Plus (55.4 vs.\ 48.2)
(Table~\ref{tab:main-results}). Diagnostics show more effective reduction
of value-estimation error as the refinement budget increases, while
ablations demonstrate that the gains are not tied to a single compaction
implementation (Sections~\ref{sec:value-refinement} and~\ref{sec:ablations}).

In summary, our contributions are:
\begin{itemize}
    \item We cast the combination of critic predictions and continuation
    evidence as an active critic learning problem (Sections~\ref{sec:method-credit} and~\ref{sec:method-search}).
    \item We introduce FOCUS, a highly customizable rollout controller,
    which allocates a limited rollout budget according to estimated
    value-estimation error reduction and integrates the resulting
    estimates with credit assignment.
    \item We show that FOCUS improves over critic-free baselines on
    long-horizon benchmarks (Section~\ref{sec:experiments}).
\end{itemize}

\section{Method}
\label{sec:method}

\subsection{Problem Setup}
\label{sec:learning-setup}

We consider reinforcement learning of a language-model policy $\pi_\theta$
from outcome rewards. The policy generates responses or actions from its
current context, and a critic $V_\phi$, parameterized by $\phi$, estimates
the expected remaining return. We assume that at least some of the intermediate states are easily forkable and that repeated sampling from them is trivial.
An episode $\xi$ records the generated turns and
rewards from one completed run.

Given a finite rollout budget, FOCUS selects from candidate states from which to sample additional continuations, refines their value estimates, and constructs
training samples for both the critic and the policy. For tool-using agents, compaction
boundaries provide natural candidate states. Appendix~\ref{app:execution} details the execution mechanics used in our experiments.

\subsection{Critic Refinement and Learning}
\label{sec:method-credit}

We first describe how collected continuations produce the learning signals for the critic and the policy.
Section~\ref{sec:method-search} describes how we allocate the
rollout budget.

\noindent\textbf{States and rollout segments.}
For each prompt $x$, we represent collected rollouts as a tree
$\mathcal T_x$. A nonterminal node denotes a state from which the policy can resume
with the same context and environment state; an edge contains the
policy-generated continuation and environment transitions until the next state.
We write $\bar c$ for the context of a node (state), leaving its associated environment
state implicit.

A completed trajectory contains nonterminal nodes
$\bar c_0=x,\bar c_1,\ldots,\bar c_M$ and a terminal node
$\bar c_{M+1}$, connected by edges
$e_i=(\bar c_{i-1}\to\bar c_i)$, $i=1,\ldots,M+1$.

\noindent\textbf{Refining state values.}
Let $V^\pi(\bar c)$ denote the expected discounted return of a continuation
from $\bar c$ under the collection policy. The critic $V_\phi(\bar c)$
predicts this value from the state's context. However, this prediction can have high variance.
We refine its prediction using the outcome evidence from the
$n(\bar c)$ continuations $\mathrm{ch}(\bar c)$ that have been expanded from
it, backed up one level as in
TreeRPO~\citep{yang2025treerpo}:
\begin{equation}
\bar G(\bar c) \;=\; \frac{1}{n(\bar c)} \sum_{\bar c' \in \mathrm{ch}(\bar c)}
\Big( R_{\bar c \to \bar c'} + \gamma^{d_{\bar c \to \bar c'}}\, \tilde V(\bar c') \Big),
\qquad \tilde V(\bar c') = 0 \text{ if } \bar c' \text{ is terminal,}
\label{eq:backup}
\end{equation}
where $\tilde V(\bar c')$ is the continuation's own refined value
(Eq.~\ref{eq:refined} below). We prefer this to a raw average of
descendant outcomes~\citep{li2025treepo} because the allocator chooses
which nodes to expand, so the leaves beneath $\bar c$ are no longer
an on-policy draw from $\bar c$; the continuation's refined value carries its
evidence back up instead, one level at a time.
We combine the critic prediction and the recursive backup with a fixed
pseudo-observation weight $\kappa$,
\begin{equation}
\tilde V(\bar c) \;=\; \frac{\kappa\, V_\phi(\bar c) + n(\bar c)\,\bar G(\bar c)}{\kappa + n(\bar c)}.
\label{eq:refined}
\end{equation}
For $n(\bar c)=0$, we set $\tilde V(\bar c)=V_\phi(\bar c)$ without
evaluating $\bar G$; terminal values are zero. As $n(\bar c)$ increases, the
blend places more weight on the backed-up mean~\citep{silver2016alphago,
tesauro2010bayesianmcts}.

\noindent\textbf{Critic supervision and actor advantages.}
After collection and the final backup, we train the critic on the recursive
target $\bar G$ at every non-terminal state:
\begin{equation}
\mathcal{L}_{\mathrm{critic}}(\phi) \;=\;
\mathbb{E}\Big[\sum_{\bar c \in \mathcal{T}_x:\; n(\bar c) > 0}
\big(V_\phi(\bar c) - \bar G(\bar c)\big)^2\Big].
\label{eq:critic}
\end{equation}
The target is held fixed during training. This update transfers the
collected continuation evidence into the critic used in subsequent training.
Within the current iteration, the refined values $\tilde V$ supply the
segment-level temporal-difference residual\footnote{Unless otherwise specified, all experiments use $\gamma=1$, $\lambda=0$, and $\kappa=1$.}
\begin{equation}
\hat A_i = \delta_i \;=\; R_i + \gamma^{d_i}\tilde V(\bar c_i)
-\tilde V(\bar c_{i-1}).
\label{eq:td}
\end{equation}
We apply each edge's detached advantage to its policy-generated task and
context-management tokens, denoted $y_{e_i}$, masking environment
observations and conditioning text. With $\pi_{\mathrm{old}}$ the
collection policy, $\rho_t$ the token-level importance ratio, and $\epsilon$
the clipping threshold, the actor minimizes
\begin{equation}
\mathcal{L}_{\mathrm{actor}}(\theta) \;=\;
-\,\mathbb{E}\Big[\sum_{i}\;\sum_{t \in y_{e_i}}
\min\big(\rho_t \hat A_i,\; \mathrm{clip}(\rho_t, 1-\epsilon, 1+\epsilon)\,\hat A_i\big)\Big],
\qquad \rho_t = \frac{\pi_\theta(y_t \mid y_{<t})}{\pi_{\mathrm{old}}(y_t \mid y_{<t})}.
\label{eq:actor}
\end{equation}
Appendix~\ref{app:actor-objective} gives the general TD($\lambda$) extension.

\subsection{Selective Rollout Allocation for Critic Learning}
\label{sec:method-search}

Branching continuations from candidate states supplies additional
value evidence~\citep{kazemnejad2024vineppo,hou2025treerl,li2025treepo,guo2025spo},
but uniform allocation does not account for differences in critic error or
the expected benefit of another sample. Given an explored tree
$\mathcal T_x$ and an additional rollout budget $B$, FOCUS
chooses which candidate node receives the next continuation. The aim is to acquire
evidence where it is expected to reduce value estimation error the most~\citep{roy2001errorreduction,mussmann2022eer}.

\noindent\textbf{Outcome evidence for selection.}
While selection uses the same formula for mixing observed continuation outcomes and critic predictions, it does not use the continuation-refined backups of its children as in
Eq.~\ref{eq:backup}.
Specifically, let $N(\bar c)$ be the number of continuation outcomes
and $\bar G_{\mathrm{obs}}(\bar c)$ their mean.
A completed continuation contributes one outcome at its starting node and at each newly created node along
that continuation. Later branching from a descendant does not add more observations
to its strict ancestors. Thus $N(\bar c)$ counts observed continuation outcomes, whereas $n(\bar c)$ in Eq.~\ref{eq:backup} counts direct children.

For binary continuation outcomes, the selection model uses a Beta prior of strength
$\kappa$ centered on $v=V_\phi(\bar c)$. Its posterior mean and expected
squared critic error are
\begin{equation}
\mu_{\mathrm{obs}} =
\frac{\kappa v + N\bar G_{\mathrm{obs}}}{\kappa+N},
\qquad
\widehat{e^2} = (v-\mu_{\mathrm{obs}})^2
+ \frac{\mu_{\mathrm{obs}}(1-\mu_{\mathrm{obs}})}{\kappa+N+1}.
\label{eq:selection-error}
\end{equation}
Notice how $\mu_{\mathrm{obs}}$ is distinct from the continuation-refined
training value $\tilde V$. The predictive variance of the next outcome is
$\widehat{\sigma^2}=\mu_{\mathrm{obs}}(1-\mu_{\mathrm{obs}})$. In particular, we
rely on $\mu_{\mathrm{obs}}$ to conduct the selection due to the observation outlined in Appendix~\ref{app:selection-rationale}.

\noindent\textbf{Expected benefit of a continuation.}
We score the expected utility of a continuation using the marginal MSE reduction of value estimation error at that candidate state.
For critic squared error $e^2$ and outcome variance $\sigma^2$, the acquisition function becomes
$\mathrm{MSE}(N)=(\kappa^2 e^2+N\sigma^2)/(\kappa+N)^2$.
Differentiating with $e^2$ and $\sigma^2$ held fixed, then substituting
the posterior estimates gives
\begin{equation}
g(\bar c) \;=\; \frac{2\kappa^{2}\widehat{e^2} + (N-\kappa)\,\widehat{\sigma^2}}
{(\kappa+N)^{3}},
\label{eq:gain}
\end{equation}
where $N=N(\bar c)$ counts completed continuation outcomes.

\begin{algorithm}[t]
\caption{Selective rollout allocation and value refinement, with continuation helper}
\label{alg:refine}
\small
\algrenewcommand{\algorithmiccomment}[1]{\hfill\(\triangleright\) {\scriptsize #1}}
\begin{algorithmic}
\Require tree $\mathcal T_x$; fixed actor $\pi_\theta$ and critic $V_\phi$; budget $B$; prior strength $\kappa$
\Require environment $\mathcal E$; branch predicate $\mathcal B$; compaction trigger $\mathcal G$ and operator $\mathcal C_\pi$
\Ensure tree $\mathcal T_x$ with critic targets $\bar G$ and refined values $\tilde V$
\State initialize $N,\bar G_{\mathrm{obs}}$ from completed suffixes; refresh $\mathcal F,g$
\algblocktitle{allocation}{\S\ref{sec:method-search}: Selective rollout allocation}
\While{$B>0\ \land\ \mathcal F\ne\emptyset$}
  \State $\bar c\gets\arg\max_{u\in\mathcal F}g(u)$; $\xi\gets\textsc{Continue}(\bar c,\pi_\theta)$
  \If{$\xi=\bot$}
    \State $B\gets B-1$ \Comment{execution failure}
  \Else
    \State $\mathcal T_x\gets\textsc{Attach}(\mathcal T_x,\bar c,\xi)$
    \State $B\gets B-d(\xi)$ \Comment{$d(\xi)$: new task transitions}
    \State $N(u),\bar G_{\mathrm{obs}}(u)\gets0$ for $u\in\mathrm{New}(\xi)$ \Comment{new nodes}
    \State $\textsc{Score}(\mathcal T_x,V_\phi)$ \Comment{cache unscored nonterminal values}
    \For{$u\in\{\bar c\}\cup\mathrm{New}(\xi)$} \Comment{$G_\xi(u)$: observed suffix return}
      \State $\bar G_{\mathrm{obs}}(u)\gets\dfrac{N(u)\bar G_{\mathrm{obs}}(u)+G_\xi(u)}{N(u)+1}$; $N(u)\gets N(u)+1$
    \EndFor
  \EndIf
  \State refresh $\mathcal F,g$
\EndWhile\algblockend{allocation}
\Statex
\algblocktitle{refinement}{\S\ref{sec:method-credit}: Value refinement and critic targets}
\State $\tilde V(u)\gets0$ for terminal nodes; $\tilde V(u)\gets V_\phi(u)$ for nonterminal leaves
\For{$u\in\mathcal T_x$ with $n(u)>0$, in reverse topological order}
  \State $\bar G(u)\gets\dfrac{1}{n(u)}\displaystyle\sum_{v\in\mathrm{ch}(u)}\bigl(R_{u\to v}+\gamma^{d_{u\to v}}\tilde V(v)\bigr)$
  \State $\tilde V(u)\gets\dfrac{\kappa V_\phi(u)+n(u)\bar G(u)}{\kappa+n(u)}$
\EndFor\algblockend{refinement}
\Statex
\State \Return $\mathcal T_x$
\Statex
\Function{Continue}{$\bar c,\pi$}
  \State $(c,s)\gets\textsc{Fork}(\bar c)$; $\xi\gets[\,]$; $\zeta\gets[\,]$ \Comment{$\zeta$: current segment}
  \While{$\neg\textsc{Terminal}(s)$}
    \algblocktitle[3em]{compaction}{Modular context management}
    \If{$\mathcal G(c)=1$}
      \State $c\gets\mathcal C_\pi(c)$\algblockend{compaction}
    \Statex
    \Else
      \State $\tau\sim\pi(\cdot\mid c)$; $(s,o,r)\sim\mathcal E(s,a(\tau))$; $c\gets c\,\|\,\tau\,\|\,o$
    \EndIf
    \State $\eta\gets\textsc{Trace}()$; $\zeta\gets\zeta\,\|\,\eta$ \Comment{task/compaction turns, rewards, transitions}
    \State $s\gets\textsc{Update}(s,\eta)$ \Comment{counters and termination flags}
    \State \textbf{if} $\textsc{Failed}(s)$ \textbf{then return} $\bot$
    \If{$\textsc{Terminal}(s)\ \lor\ \mathcal B(c,s)=1$} \Comment{$\mathcal B$ independent of $\mathcal G$}
      \State $\bar c'\gets c$; $\xi\gets\xi\,\|\,[(\bar c',\textsc{Snapshot}(s),\zeta)]$; $\zeta\gets[\,]$ \Comment{save resumption state}
    \EndIf
  \EndWhile
  \State \Return $\xi$
\EndFunction
\end{algorithmic}
\algblockbox{allocation}{AlgAllocation}
\algblockbox{refinement}{AlgRefinement}
\algblockbox{compaction}{AlgContext}
\end{algorithm}

Algorithm~\ref{alg:refine} first acquires rollout evidence
(Section~\ref{sec:method-search}), then refines values bottom-up
(Section~\ref{sec:method-credit}). Each continuation completes before the
next allocation decision. The eligible set is
$\mathcal F=\{\bar c\in\mathcal T_x:\bar c\text{ nonterminal and resumable},\,
 g(\bar c)>0\}$.
The included $\textsc{Continue}(\bar c,\pi)$ helper returns new nodes and
connecting segments through termination. Since $g$ uses observed suffix
statistics and fixed critic predictions, value refinement can follow collection.
Appendix~\ref{app:rl-loop} gives the full training loop.

\noindent\textbf{Practical considerations.}
Total branching width at any node is uncapped. The allocation phase can
also run asynchronously, with each completed continuation updating the
acquisition statistics while others remain in flight. In this case,
we replace $N$ in Eq.~\ref{eq:gain} by
$N_{\mathrm{eff}}(\bar c)=N(\bar c)+m(\bar c)$, where $m$ counts pending
continuations; posterior estimates still use completed outcomes only.
We limit concurrent continuations per node to $C=2$ and globally to
worker capacity $W$, restricting eligibility to $m(\bar c)<C$.
Completed children do not consume concurrency slots. In practice, the two
loops can run concurrently, each with multiprocessing. After the launch
budget is exhausted, pending continuations finish and a final bottom-up
refinement ensures that the training targets reflect the completed tree.

\section{Experiments}
\label{sec:experiments}

We compare FOCUS to GRPO and VinePPO on three long-horizon benchmarks,
using Qwen3.5-9B on all three and Qwen3.5-27B~\citep{qwen2026qwen35} (LoRA) on
DeonticBench. We assess how MCTS refinement
improves critic estimates and ablate the compaction operator and refinement rollouts.
Appendix~\ref{sec:math-sanity} reports a preliminary comparison with PPO and GRPO
on MATH using prefix continuations without compaction.

\subsection{Long-horizon datasets}
\noindent\textbf{DeonticBench \citep{dou2026deonticbench}}
DeonticBench evaluates models on context-grounded deontic reasoning:
deriving obligations, permissions, and prohibitions from real statutes
and case facts.
It comprises 6,232 tasks across U.S.\ federal tax, airline baggage
policy, immigration administration, and state housing law.
Applying a long statute to a specific case is a multi-step process,
which makes the benchmark a setting for long-horizon agentic reasoning
in which the model consults the statutes on demand~\citep{dou2026dar}
and a natural testbed for stepwise rollout. We focus on the \textsc{airline},
\textsc{sara-numeric}, and \textsc{sara-binary} subsets, whose statutes are
reasonably long and involve non-trivial long-horizon tasks. The actor harness
exposes \tool{list}, \tool{grep}, \tool{read}, and \tool{python}.
We stratify the data into equally sized training and held-out splits.

\noindent\textbf{BrowseComp-Plus \citep{chen-etal-2026-browsecomp}}
BrowseComp-Plus evaluates deep search agents on complex, multi-hop
information-seeking queries over a fixed, human-verified corpus rather
than a live web API. The resulting trajectories are long-horizon: iterative query
formulation, retrieval, and reflection, which makes the benchmark a
natural testbed for stepwise rollout. Following TRACE~\citep{tao2026trace}, the
actor harness exposes \tool{search}, \tool{open}, and \tool{find} over the
fixed corpus. We randomly split the 830 benchmark test tasks into 498 training,
83 validation, and 249 held-out test tasks. For each method, we select the
checkpoint with the highest accuracy on the 83-task validation split and
report its accuracy on the 249-task held-out test split.

\noindent\textbf{LocBench \citep{chen2025locagent}}
LocBench is a code-localization benchmark of 560 real GitHub issues
(242 bug reports, 150 feature requests, 139 performance issues, and 29
security issues) from repositories such as \texttt{transformers},
\texttt{scikit-learn}, \texttt{vllm}, and \texttt{ultralytics}; in
LocAgent's construction, bug reports are issues created relatively recently, and security and performance instances are collected via
GitHub keyword search. Given only the issue,
the agent must locate the files and functions touched by the gold patch.
We instantiate it using the harness described in Appendix~\ref{app:execution}; the actor
exposes \tool{list}, \tool{grep}, and \tool{read} for repository
navigation. We score submissions as in LocAgent: file-level and function-level Acc@$k$ is~$1$
when every gold location appears in the top~$k$ (added functions are
excluded); we report Acc@$\{1,5\}$ at the file level and train on
file-level recall at $k{=}5$.

\subsection{Main results}

We compare GRPO, VinePPO, and FOCUS using Qwen3.5-9B on DeonticBench, BrowseComp-Plus, and LocBench, and additionally with Qwen3.5-27B (LoRA) on DeonticBench.
Table~\ref{tab:main-results} reports held-out accuracy alongside untrained
reference models. All methods train for one epoch under a matched per-step
generation budget with compaction on every rollout; protocol, optimization, and
compute details are in Appendix~\ref{app:training-setup}.

\newcommand{\deltaNine}{%
\multicolumn{2}{r}{$\Delta$ vs. GRPO} & $+7.3_{+0.67}^{+14.00}$ & $-4.0_{-16.00}^{+8.00}$ & $+3.6_{+0.00}^{+7.25}$ & $-1.4_{-8.45}^{+5.63}$ & $+4.1_{+0.59}^{+7.69}$ & $+7.2_{+2.41}^{+12.05}$ & $+2.3_{-1.15}^{+8.05}$ & $+0.0_{-2.30}^{+2.30}$ \\
}
\newcommand{\deltaTwentySeven}{%
\multicolumn{2}{r}{$\Delta$ vs. GRPO} & $+2.0_{-4.67}^{+8.67}$ & $+2.0_{-6.00}^{+10.00}$ & $+2.2_{+0.72}^{+4.35}$ & $+0.0_{-7.04}^{+7.04}$ & $+2.0_{-1.18}^{+5.33}$ & \omitbench \\
}
\begin{table}[htbp]
\caption{Evaluation results for FOCUS vs critic-free methods. Entries are accuracy~(\%) on the held-out split of
each benchmark. DeonticBench Hard aggregates dataset-marked hard subsets from Airline, SARA-N, and SARA-B.
Reference rows are untrained. $\Delta$ is FOCUS minus GRPO in percentage points;
subscripts and superscripts give the lower and upper endpoints of marginal
90\% paired bootstrap CIs, respectively.
For cost reasons, we only report DeonticBench at Qwen3.5-27B.}
\label{tab:main-results}
\centering
\setlength{\tabcolsep}{3pt}
\footnotesize
\resizebox{\linewidth}{!}{%
\begin{tabular}{lrcccccc>{}c>{}c}
\toprule
\multirow{2}{*}{Model} & \multirow{2}{*}{Method}
  & \multicolumn{5}{c}{DeonticBench}
  & \multirow{2}{*}{\shortstack{BrowseComp\\-Plus}}
  & \multicolumn{2}{c}{LocBench} \\
\cmidrule(lr){3-7}\cmidrule(lr){9-10}
& & Airline & SARA-N & SARA-B & Hard & Overall & & File@1 & File@5 \\
\midrule
\multicolumn{2}{l}{\emph{Reference (no RL)}} & & & & & & & & \\
\multicolumn{2}{l}{gpt-oss-20b (medium)} & 2.7 & 8.0 & 76.8 & 14.1 & 33.7 & 18.6 & 54.0 & 57.5 \\
\multicolumn{2}{l}{Nemotron~3 Nano 30B A3B}  &  2.7 & 12.0 & 89.9 & 18.3 & 36.1 & 19.3 & 57.9 & 66.6 \\
\multicolumn{2}{l}{Qwen3.5-9B}              & 5.3 & 30.0 & 89.9 & 19.7 & 45.9 & 29.7 & 62.1 & 67.8 \\
\multicolumn{2}{l}{Qwen3.5-27B}             & 58.7 & 60.0 & 94.2 & 43.7 & 73.4 & 50.6 & 66.7 & 79.3 \\
\midrule
\multirow{3}{*}{Qwen3.5-9B}
  & GRPO                     & 14.0 & \textbf{56.0} & 92.8 & \textbf{28.2} & 52.4 & 48.2 & 70.1 & \textbf{80.5} \\
  & VinePPO                  & 10.0 & 50.0 & 92.0 & 24.0 & 49.4 & 46.6 & 67.8 & 79.3 \\
  & \textbf{FOCUS} & \textbf{21.3} & 52.0 & \textbf{96.4} & 26.8 & \textbf{56.5} & \textbf{55.4} & \textbf{72.4} & \textbf{80.5} \\
\addlinespace
\deltaNine
\midrule
\multirow{3}{*}{Qwen3.5-27B}
  & GRPO                     & 60.7 & 60.0 & 94.2 & \textbf{43.7} & 74.3 & \omitbench \\
  & VinePPO                  & \textbf{62.7} & 58.0 & 94.2 & 36.6 & 74.9 & \omitbenchlabel{outside the 27B study} \\
  & \textbf{FOCUS} & \textbf{62.7} & \textbf{62.0} & \textbf{96.4} & \textbf{43.7} & \textbf{76.3} & \omitbench \\
\deltaTwentySeven
\bottomrule
\end{tabular}}
\end{table}

At Qwen3.5-9B, FOCUS leads on DeonticBench overall in Table~\ref{tab:main-results}
(56.5 vs.\ 52.4 for GRPO and 49.4 for VinePPO), reports the strongest
BrowseComp-Plus held-out test result (55.4 vs.\ 48.2 for GRPO and 46.6 for
VinePPO, using each method's validation-selected checkpoint). On LocBench, FOCUS leads on File@1
and ties GRPO on File@5. FOCUS also leads
across scale: Qwen3.5-27B with LoRA on DeonticBench still gains from
FOCUS fine-tuning (76.3 overall vs.\ 74.3 and 74.9), but the margin over the
baselines shrinks. This could partially be because the untuned 27B model already starts much higher
(73.4 overall).

\begin{figure}[!t]
\centering
\begin{subfigure}{0.32\linewidth}
  \centering
  \includegraphics[width=\linewidth]{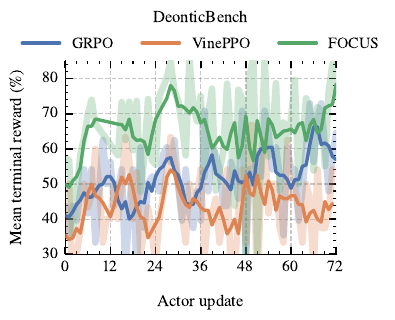}\\[3pt]
  \includegraphics[width=\linewidth]{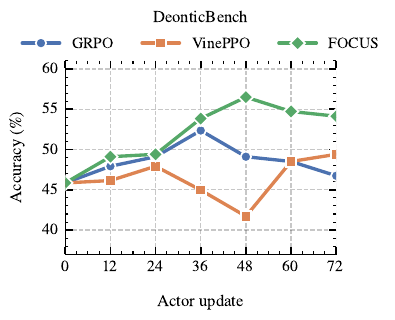}\\[3pt]
  \includegraphics[width=\linewidth]{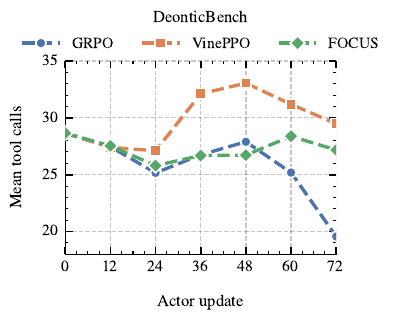}
  \caption{DeonticBench.}
  \label{fig:curves-deontic}
\end{subfigure}\hfill
\begin{subfigure}{0.32\linewidth}
  \centering
  \includegraphics[width=\linewidth]{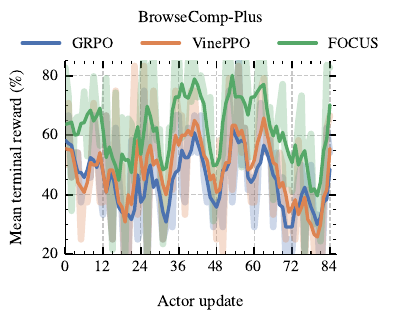}\\[3pt]
  \includegraphics[width=\linewidth]{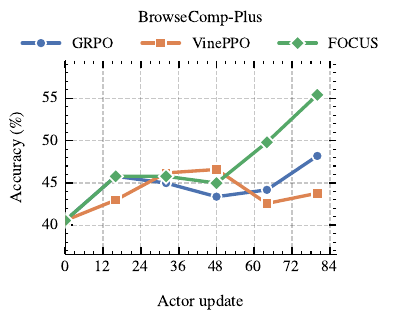}\\[3pt]
  \includegraphics[width=\linewidth]{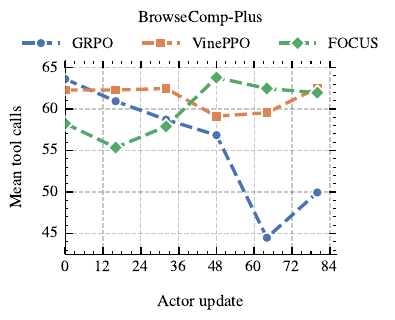}\\[3pt]
  \caption{BrowseComp-Plus.}
  \label{fig:curves-bcp}
\end{subfigure}\hfill
\begin{subfigure}{0.32\linewidth}
  \centering
  \includegraphics[width=\linewidth]{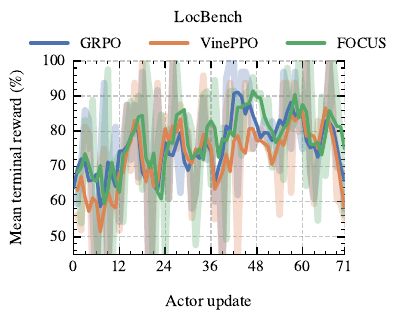}\\[3pt]
  \includegraphics[width=\linewidth]{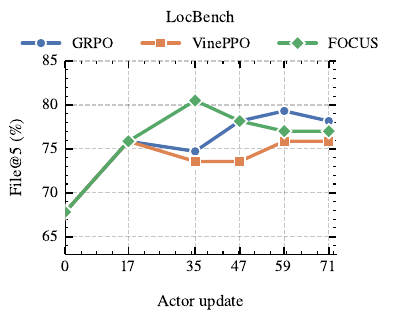}\\[3pt]
  \includegraphics[width=\linewidth]{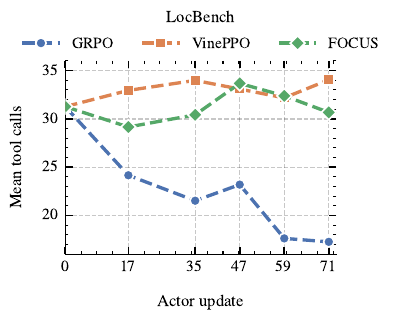}

  \caption{LocBench.}
  \label{fig:curves-loc}
\end{subfigure}
\caption{Each column is one benchmark, showing training
reward against optimizer step (top), held-out evaluation (middle), and mean numbers of tool calls per episode on the held-out
split (bottom), for GRPO, VinePPO, and FOCUS under a matched
per-step generation budget. Top panels plot mean terminal reward from
on-policy collection; LocBench held-out reward is File@5. BrowseComp-Plus results are the equivalence judgement of Qwen3-32B~\citep{qwen2025qwen3}.}
\label{fig:curves}
\end{figure}

\noindent\textbf{Training dynamics.}
FOCUS attains the highest on-policy reward throughout training on DeonticBench
and BrowseComp-Plus (Figure~\ref{fig:curves}), and the highest held-out
accuracy on DeonticBench. On DeonticBench and LocBench, the three methods track
closely early, then FOCUS pulls ahead and keeps improving or holds a
higher plateau, whereas GRPO on DeonticBench peaks mid-run and then regresses.
Late in training, GRPO's mean tool calls collapse on all three benchmarks while
VinePPO and FOCUS stay near their initial levels: GRPO increasingly
\tool{submit}s right after compaction even when the statute or repository still
needs reading (Appendix~\ref{app:grpo-rollouts}). We hypothesize that
trajectory-level group credit makes such premature submissions easy to
reinforce, whereas VinePPO's suffix branching and FOCUS's refined checkpoint
values preserve longer horizons.

\subsection{Value refinement}
\label{sec:value-refinement}

To assess how efficient the FOCUS correction process is in reducing the critic error for long-horizon tasks, as well as whether the critic
can improve throughout training, we conduct two diagnostic analyses.

\begin{figure}[htbp]
  \centering
  \begin{subfigure}[t]{0.48\linewidth}
    \centering
    \includegraphics[width=\linewidth]{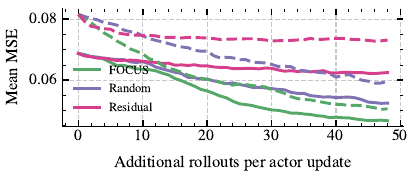}
    \caption{Critic MSE vs.\ budget (DeonticBench).\label{fig:critic-mse-budget}}
  \end{subfigure}\hfill
  \begin{subfigure}[t]{0.48\linewidth}
    \centering
    \includegraphics[width=\linewidth]{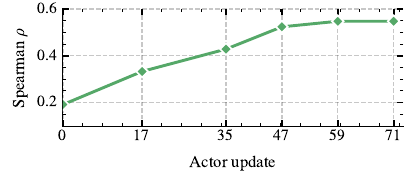}
    \caption{Critic Spearman $\rho$ (LocBench).\label{fig:critic-pearson-locbench}}
  \end{subfigure}
  \caption{Value refinement for FOCUS. (a)~MSE on fixed audit trees as MCTS
  continuations are allocated by each rule (solid: constant critic init;
  dashed: random critic init). (b)~Rank agreement between checkpoint critic
  values and an oversampled, critic-blind MCTS target on fixed audit states.}
  \label{fig:value-refinement}
\end{figure}

\noindent\textbf{Critic Error Reduction} We start by assuming two types of imperfect critics:
a constant critic init and a random critic init, and we compare how different priority
schedules affect the critic error reduction against a ground truth determined by the mean outcome of $n{=}16$
MCTS continuations. \autoref{fig:critic-mse-budget} shows that under both types of critics, FOCUS's correction process
is noticeably more efficient than random selection. Also, using the raw critic value residual performs poorly.
This shows that FOCUS-style scheduling not only works well under degenerate critics but also reduces MSE even when the critic provides almost no information.

\noindent\textbf{Critic Improvement} We also test whether the critic improves over training.
At each LocBench evaluation checkpoint, we sample $n{=}16$ MCTS continuations from a set of $m{=}8$ completed traces.
We then compute the Spearman correlation between the critic value and the MCTS ground truth. \autoref{fig:critic-pearson-locbench} shows that the FOCUS critic improves steadily.

\subsection{Ablation studies}
\label{sec:ablations}

We study how rollout allocation, compaction operators, refinement rollouts, compaction prompts and threshold impact FOCUS performance.

\begingroup
\noindent{\textbf{Rollout allocation.}}
We additionally evaluate two budget allocation strategies on DeonticBench while keeping the remaining
FOCUS configuration fixed. Random allocates additional continuations through
uniformly sampling from any existing intermediate state, and TRACE~\citep{zou2026trace}, which
allocates continuations according to how likely they will lead to mixed outcomes. For TRACE, we
adopt the Active Stage 2 configuration. Table~\ref{tab:allocation-ablation}
shows how both strategy compare with FOCUS. While FOCUS outperforms both on overall
DeonticBench performance, the gap with TRACE is smaller. Appendix~\ref{app:trace-relationship}
shows that their initial single-continuation priority rankings coincide under binary
rewards when both rules use the same predictions and FOCUS uses $\kappa=1$.
\par\endgroup

\begin{table}[htbp]

\arrayrulecolor{black}
\centering
\caption{Rollout-allocation comparison on DeonticBench.}
\label{tab:allocation-ablation}
\begingroup
\begin{tabular}{lrrrrr}
\toprule
Method & Airline & SARA-N & SARA-B & Hard & Overall \\
\midrule
Random allocation & 16.7 & 46.0 & 94.2 & 23.9 & 52.7 \\
TRACE & 27.3 & 42.0 & 92.0 & 29.6 & 55.9 \\
FOCUS & 21.3 & 52.0 & 96.4 & 26.8 & 56.5 \\
\bottomrule
\end{tabular}
\endgroup
\arrayrulecolor{black}
\end{table}

\begin{figure}[htbp]
\centering
\begin{subfigure}[t]{0.24\linewidth}
  \centering
  \includegraphics[width=\linewidth]{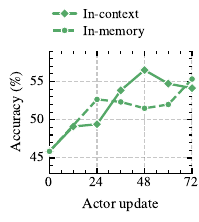}
  \caption{Compaction operator (DeonticBench).}
  \label{fig:ablation-compactor}
\end{subfigure}\hfill
\begin{subfigure}[t]{0.24\linewidth}
  \centering
  \includegraphics[width=\linewidth]{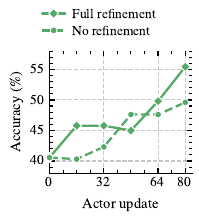}
  \caption{Refinement rollouts (BrowseComp-Plus).}
  \label{fig:ablation-active-learning}
\end{subfigure}\hfill
\begin{subfigure}[t]{0.24\linewidth}
  \centering
  \includegraphics[width=\linewidth]{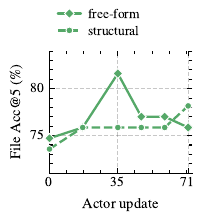}
  \caption{Structural compaction (LocBench).}
  \label{fig:ablation-prompt}
\end{subfigure}\hfill
\begin{subfigure}[t]{0.24\linewidth}
  \centering
  \includegraphics[width=\linewidth]{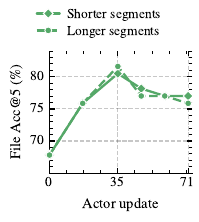}

  \caption{Compaction threshold (LocBench).}
  \label{fig:ablation-threshold}
\end{subfigure}
\caption{Ablations at evaluation checkpoints. (a)~Varying the compaction
operator on DeonticBench under a matched budget; (b)~FOCUS with full
refinement vs.\ FOCUS without refinement rollouts on BrowseComp-Plus;
(c)~free-form vs.\ structural compaction on LocBench; (d)~LocBench configurations with 24,576- and 32,768-token compaction
thresholds.}
\label{fig:ablations}
\end{figure}

\noindent\textbf{Compaction operator.}
On DeonticBench, we replace the in-context compaction operator, which writes everything into the new
context window, with an in-memory workspace that provides its own \tool{put}, \tool{get}, \tool{scan}
and \tool{delete}. The two perform similarly (Figure~\ref{fig:ablation-compactor}), so FOCUS's
effectiveness extends to different compaction operators.

\noindent\textbf{Refinement rollouts.} On BrowseComp-Plus, we remove the refinement rollouts of \autoref{sec:method-search},
which amounts to collecting independent continuations from the question prompt and estimating step-wise
advantages from the critic alone. Figure~\ref{fig:ablation-active-learning} shows
that refinement rollouts are crucial for FOCUS on BrowseComp-Plus; without them, FOCUS trains
more slowly, especially early on.

\noindent\textbf{Structural compaction.}
We also compare a free-form compaction prompt with a structural prompt
that induces a more constrained compaction on LocBench
(Figure~\ref{fig:ablation-prompt}; both prompts are reproduced in
Appendix~\ref{app:compaction-prompts}).
The free-form prompt suggests headings and keeps the reply verbatim. The
structural prompt requires a single summary with four sections (State,
Evidence, Open questions, Next steps) and rejects replies that do not parse.
The free-form prompt reaches a higher peak, while the structural prompt improves more steadily.

\begingroup
\noindent\textbf{Compaction threshold.}
While the main configuration on LocBench uses a 24,576-token compaction
threshold, we compare that with a longer 32,768-token
threshold (Figure~\ref{fig:ablation-threshold}). Best performing File@1 is 72.4\% in both
configurations, while File@5 is 80.5\% and 81.6\%, respectively. The evaluation results closely tracks each other across steps.
\par\endgroup

\section{Related Work}
\label{sec:related}

\noindent\textbf{Structured rollouts and rollout allocation}
Search over intermediate states began as an inference-time technique:
LLM tree search~\citep{yao2023tot,hao2023rap,zhou2023lats} and verifier-guided
test-time scaling~\citep{lightman2024verify,snell2024scaling}
improved model responses. It was later used offline to generate training supervision~\citep{tian2024alphallm,
zhang2024restmcts,guan2025rstarmath} and then moved into the RL loop.
VinePPO~\citep{kazemnejad2024vineppo}, SPO~\citep{guo2025spo}, and TEMPO~\citep{tran2025tempo} re-sample continuations from
intermediate reasoning steps to replace the value network with Monte Carlo estimates; TreeRL~\citep{hou2025treerl}, TreePO~\citep{li2025treepo},
and DeepSearch~\citep{wu2025deepsearch} run the search inside RLVR training
so that shared prefixes cut generation cost and the tree yields
process-level signal. Other work proposes where to branch: at
high-entropy tokens~\citep{zheng2025fr3e,cao2026treeadv}, at attention
peaks~\citep{liu2025attnrl}, where the model's belief
shifts~\citep{lei2026beliefshift}, or, for tool-using agents, at the turn
boundaries after an observation~\citep{dong2025arpo,zong2026at2po}. Tree-GRPO~\citep{ji2025treegrpo} makes each agent step a node and
shows that intra-tree group objectives amount to step-level preference
learning, while PaTR~\citep{li2026patr} routes the branching budget by a
process score, and IGRPO~\citep{zhang2026igrpo} and
SIGHT~\citep{zhong2026sight} by information gain. T$^2$PO~\citep{wang2026t2po}
instead resamples turns that its token- and turn-level uncertainty signals
flag as making no progress. TRACE~\citep{zou2026trace} unifies prompt- and
turn-level allocation with a learned predictor of conditional success,
directing extra rollouts to the prefixes most likely to yield mixed
terminal rewards.
FOCUS allocates additional continuations according to their estimated reduction
in value-estimation error, using the acquired evidence to refine state values
and train a reusable critic.

\noindent\textbf{Active learning and expected error reduction}
Expected-error-reduction methods select observations according to their
predicted effect on the learner's error. \citet{roy2001errorreduction}
approximate this objective through sampling, while
\citet{mussmann2022eer} use Bayesian parameter samples to estimate the
reduction in predictive loss over an unlabeled validation set without
retraining for every candidate query. A related approach selects data to
reduce predictive variance, connecting variance reduction to squared-error
minimization under an approximately unbiased learner~\citep{cohn1996active}.
Influence-function approaches estimate acquisition utility by approximating
how additional training data affect a fitted model's evaluation loss,
without retraining for each candidate~\citep{koh2017influence,xu2019goaloriented}.
FOCUS applies the error-reduction principle to allocating continuation
samples among intermediate states.

\noindent\textbf{Critic learning and credit assignment}
Trajectory-level reward is known to fail for multi-turn agents as it cannot
tell a useful turn from a redundant one, and training on it alone is
unstable~\citep{wang2025ragen,wei2025mtgrpo,tan2026hcapo}. One response is
to place the reward and critic at the level of a turn or
step rather than a token (too fine-grained) or a whole
trajectory (too coarse)~\citep{zhou2024archer,zhou2025sweetrl,li2026turnppo,
zhang2026hygae,zhen2026stephrl}. A turn is the natural unit of decision
for an agent; the horizon measured in turns is short enough for a critic to
bootstrap over, and a per-turn value gives every turn its own signal from
a single rollout. Large-scale training is also
shifting from structural comparison to a learned value: SAO~\citep{hou2026sao}
replaces GRPO's group sampling with one rollout per prompt scored by a
value model, as full groups are slow and off-policy under asynchronous training.
Some process reward models serve as fixed value estimators during policy optimization~\citep{wang2023mathshepherd,luo2024omegaprm,li2024pqm,choudhury2025agentprm,setlur2024pav,tao2026trace}. More recently, Le Critique~\citep{venkatraman2026lecritique} mixes a learned baseline with a training-free group-relative one according to how accurate the critic is.
FOCUS uses refined state values to construct segment-level actor advantages
and recursive backups to supervise the critic, coupling credit assignment
in the current iteration with critic learning for subsequent iterations.

\noindent\textbf{Context management for long-horizon agents}
Compaction is becoming standard for agents that outrun their context
window~\citep{packer2023memgpt,yu2026memagent,sun2025contextfolding, li2026self}, and
there is a recent push to train it jointly with the task rather than treat
it as a frozen preprocessor~\citep{cursor2026selfsummarization,
li2026compactionrl}. State restoration is also central to identifying step-level credit
\citep{zhang2026creditsurvey}. In our long-horizon experiments, compaction
boundaries provide resumable states for acquiring continuation evidence.
The rollout controller operates on these states independently of the
compaction operator; its objective is to refine value estimates and
construct critic supervision through selective rollouts.

\section{Conclusion}
\label{sec:conclusion}

We present FOCUS, a rollout controller for language-model reinforcement
learning that refines intermediate value estimates through active critic
learning and selective rollouts. FOCUS naturally supports long-horizon
agentic RL through flexible and modular context-management protocols such
as compaction.
Across DeonticBench, BrowseComp-Plus, and LocBench, FOCUS improves over GRPO
and VinePPO under a matched per-step generation budget, with the largest
held-out gains on deep search and deontic reasoning and consistent benefits
on code localization. Training curves show that these improvements come with
more stable exploration: FOCUS maintains higher collection reward and avoids
the late-training collapse in tool use seen under trajectory-level group
credit. Ablations indicate that gains are not tied to a single
compaction implementation and that refinement rollouts are beneficial on BrowseComp-Plus. We release our modular rollout controller with interfaces for
alternative harnesses, search policies, and credit estimators. Future work may scale FOCUS to
larger models and longer budgets, tighten the link between compaction
quality and training stability, and adopt stronger uncertainty
quantification for value refinement.

\subsection*{Acknowledgments}

This work was supported in part by the Defense Advanced Research Projects
Agency (DARPA) Scientific Feasibility (SciFy) program under contract
HR001125C0304; by Schmidt Sciences through the ``Safely Confident Reasoning''
project in its Science of Trustworthy AI program; and by the Office of Naval
Research (ONR) under grant N00014-24-1-2089.
Any opinions, findings, and conclusions or recommendations expressed in this
material are those of the authors and do not necessarily reflect the views of
DARPA, the Office of Naval Research, or Schmidt Sciences. We acknowledge the
use of computational resources from the Johns Hopkins DSAI cluster.

\subsection*{AI use statement}

In this work, we used generative AI tools to generate synthetic data for imitation learning to compact.
We used models from the same family as the actor model being trained to compact already-generated trajectory steps when cold start is needed, yielding pseudo-compaction traces.
We also used generative AI tools to help explore design ideas for the active rollout acquisition phase, to assist in implementing methods, and to clean and reformat datasets.
We have not used generative AI tools to develop theoretical models or conceptual frameworks, assist with translation, support qualitative or thematic data analysis, or
interpret results. AI assistance is used to formalize and then restructure the proof for Appendix I.
Additionally, we used generative AI tools to identify relevant literature and to help draft
and edit the paper for clarity and readability.

We have reviewed all AI-assisted work: generated pseudo compaction traces were inspected by the
authors, design ideas suggested by AI were evaluated and selected by the authors, AI-assisted
code was reviewed and tested by the authors before use in experiments, reformatted data was
spot-checked against the original sources, all cited works were verified by the authors, and
all AI-assisted text was reviewed and revised by the authors.
We take responsibility for the final content of this work, including text, claims, or artifacts
produced with the aid of generative AI.

\subsection*{Reproducibility statement}

The supplemental material contains the implementation of our rollout-controller
library and of the DeonticBench, BrowseComp-Plus, and LocBench environments
used in our experiments. Section~\ref{sec:method} and
Algorithm~\ref{alg:refine} specify the method, and
Appendix~\ref{app:training-setup} details the training protocol and the
hyperparameters for every run.

\clearpage
\appendix
\section{Training setup}
\label{app:training-setup}

Tables~\ref{tab:setup-dt-9b}--\ref{tab:setup-dt-27b} group the training
settings by benchmark and model size, following the order in
Table~\ref{tab:main-results}. Each table compares GRPO, VinePPO, and FOCUS;
a dash marks a method-specific setting that does not apply.
Adam is from \citet{kingma2014adam}.

\noindent\textbf{Shared FOCUS settings.}
Across all FOCUS experiments, we use $\gamma=1$, $\lambda=0$, and
$\kappa=1$. Thus rewards are undiscounted, actor advantages use the local
segment TD residual in Eq.~\ref{eq:td}, and the critic contributes one
pseudo-observation to both the recursive value blend and the selection
posterior. Total branching width is uncapped, with at most $C=2$
concurrent continuations per node; completed children do not count
against this concurrency limit. These settings are shared across tasks
and model sizes.
The selection estimators are defined in Eq.~\ref{eq:selection-error} and
Appendix~\ref{app:selection-rationale}; the full training loop and budget
semantics are given in Appendix~\ref{app:rl-loop}.

\begin{table}[htbp]
\centering
\caption{Training setup for DeonticBench with Qwen3.5-9B.}
\label{tab:setup-dt-9b}
\label{tab:setup-dt-focus}
\label{tab:setup-dt-grpo}
\label{tab:setup-dt-vineppo}
\footnotesize
\setlength{\tabcolsep}{3pt}
\begin{tabular}{@{}>{\raggedright\arraybackslash}p{0.31\linewidth}*{3}{>{\raggedright\arraybackslash}p{0.21\linewidth}}@{}}
\toprule
Setting & GRPO & VinePPO & FOCUS \\
\midrule
\multicolumn{4}{@{}l}{\textbf{Protocol}} \\
Actor model & Qwen3.5-9B & Qwen3.5-9B & Qwen3.5-9B \\
Critic model & \textemdash & \textemdash & 9B \\
Training length & 1 epoch & 1 epoch & 1 epoch \\
Training-data order & Identical across methods & Identical across methods & Identical across methods \\
Compaction operator & On & On & On \\
Rollout temperature & 1.0 & 1.0 & 1.0 \\
Advantage cutoff & 0.04 & 0.04 & 0.04 \\
Seeding budget $B_{\mathrm{seed}}$ (turns/prompt) & \textemdash & \textemdash & 64 \\
Step advantage & Group advantage broadcast over steps & \textemdash & \textemdash \\
Branching states & \textemdash & Compaction points & \textemdash \\
Value-only continuations per state & \textemdash & 1 ($k{=}1$) & \textemdash \\
\midrule
\multicolumn{4}{@{}l}{\textbf{Rollout grouping (per actor update)}} \\
Group size & 8 & 4 & Adaptive search \\
Questions & 6 & 6 & 6 \\
Root trajectories & 48 & 24 & Set by adaptive search \\
\midrule
\multicolumn{4}{@{}l}{\textbf{Harness}} \\
Compaction trigger (prompt tokens) & 14{,}336 & 14{,}336 & 14{,}336 \\
Compaction trigger (tool calls) & 18 & 18 & 18 \\
Actor reply cap (tokens) & 8{,}144 & 8{,}144 & 8{,}144 \\
Compaction reply cap (tokens) & 4{,}096 & 4{,}096 & 4{,}096 \\
Agent turn limit (turns) & 48 & 48 & 48 \\
Context window (tokens) & 32{,}768 & 32{,}768 & 32{,}768 \\
\midrule
\multicolumn{4}{@{}l}{\textbf{Optimization}} \\
Parameterization & Full & Full & Full \\
Optimizer & Adam & Adam & Adam \\
Actor learning rate & $1\times 10^{-6}$ & $1\times 10^{-6}$ & $1\times 10^{-6}$ \\
Critic learning rate & \textemdash & \textemdash & $5\times 10^{-6}$ \\
Learning-rate schedule & Constant & Constant & Constant \\
Adam $(\beta_1, \beta_2, \epsilon)$ & $(0.9,\ 0.95,\ 10^{-8})$ & $(0.9,\ 0.95,\ 10^{-8})$ & $(0.9,\ 0.95,\ 10^{-8})$ \\
Weight decay & 0 & 0 & 0 \\
Gradient-norm clip & 1 & 1 & 1 \\
PPO clip ratio & 0.2 & 0.2 & 0.2 \\
KL coefficient & 0.01 & 0.01 & 0.01 \\
Entropy bonus & 0 & 0 & 0 \\
\midrule
\multicolumn{4}{@{}l}{\textbf{Compute}} \\
Training GPUs & 2$\times$H200 & 2$\times$H200 & 2$\times$H200 \\
Rollout GPUs & 4--6$\times$H100 & 4--6$\times$H100 & 4--6$\times$H100 \\
Critic GPUs (refinement) & \textemdash & \textemdash & 1$\times$H100 \\
\bottomrule
\end{tabular}
\end{table}

\begin{table}[htbp]
\centering
\caption{Training setup for BrowseComp-Plus with Qwen3.5-9B.}
\label{tab:setup-bc-9b}
\label{tab:setup-bc-focus}
\label{tab:setup-bc-grpo}
\label{tab:setup-bc-vineppo}
\footnotesize
\setlength{\tabcolsep}{3pt}
\begin{tabular}{@{}>{\raggedright\arraybackslash}p{0.31\linewidth}*{3}{>{\raggedright\arraybackslash}p{0.21\linewidth}}@{}}
\toprule
Setting & GRPO & VinePPO & FOCUS \\
\midrule
\multicolumn{4}{@{}l}{\textbf{Protocol}} \\
Actor model & Qwen3.5-9B & Qwen3.5-9B & Qwen3.5-9B \\
Critic model & \textemdash & \textemdash & 9B \\
Training length & 1 epoch & 1 epoch & 1 epoch \\
Training-data order & Identical across methods & Identical across methods & Identical across methods \\
Compaction operator & On & On & On \\
Rollout temperature & 1.0 & 1.0 & 1.0 \\
Advantage cutoff & 0.04 & 0.04 & 0.04 \\
Seeding budget $B_{\mathrm{seed}}$ (turns/prompt) & \textemdash & \textemdash & 64 \\
Step advantage & Group advantage broadcast over steps & \textemdash & \textemdash \\
Branching states & \textemdash & Compaction points & \textemdash \\
Value-only continuations per state & \textemdash & 1 ($k{=}1$) & \textemdash \\
\midrule
\multicolumn{4}{@{}l}{\textbf{Rollout grouping (per actor update)}} \\
Group size & 8 & 4 & Adaptive search \\
Questions & 6 & 12 & 6 \\
Root trajectories & 48 & 48 & Set by adaptive search \\
\midrule
\multicolumn{4}{@{}l}{\textbf{Harness}} \\
Compaction trigger (prompt tokens) & 14{,}336 & 14{,}336 & 14{,}336 \\
Compaction trigger (tool calls) & 18 & 18 & 18 \\
Actor reply cap (tokens) & 16{,}384 & 16{,}384 & 16{,}384 \\
Compaction reply cap (tokens) & 16{,}384 & 16{,}384 & 16{,}384 \\
Agent turn limit (turns) & 80 & 80 & 80 \\
Context window (tokens) & 98{,}304 & 98{,}304 & 98{,}304 \\
\midrule
\multicolumn{4}{@{}l}{\textbf{Optimization}} \\
Parameterization & Full & Full & Full \\
Optimizer & Adam & Adam & Adam \\
Actor learning rate & $1\times 10^{-6}$ & $1\times 10^{-6}$ & $1\times 10^{-6}$ \\
Critic learning rate & \textemdash & \textemdash & $1\times 10^{-6}$ \\
Learning-rate schedule & Constant & Constant & Constant \\
Adam $(\beta_1, \beta_2, \epsilon)$ & $(0.9,\ 0.95,\ 10^{-8})$ & $(0.9,\ 0.95,\ 10^{-8})$ & $(0.9,\ 0.95,\ 10^{-8})$ \\
Weight decay & 0 & 0 & 0 \\
Gradient-norm clip & 1 & 1 & 1 \\
PPO clip ratio & 0.2 & 0.2 & 0.2 \\
KL coefficient & 0.01 & 0.01 & 0.01 \\
Entropy bonus & 0 & 0 & 0 \\
\midrule
\multicolumn{4}{@{}l}{\textbf{Compute}} \\
Training GPUs & 2$\times$H200 & 2$\times$H200 & 2$\times$H200 \\
Rollout GPUs & 4--6$\times$H100 & 4--6$\times$H100 & 4--6$\times$H100 \\
Critic GPUs (refinement) & \textemdash & \textemdash & 1$\times$H100 \\
\bottomrule
\end{tabular}
\end{table}

\begin{table}[htbp]

\arrayrulecolor{black}
\centering
\caption{Training setup for LocBench with Qwen3.5-9B.}
\label{tab:setup-lb-9b}
\label{tab:setup-lb-focus}
\label{tab:setup-lb-grpo}
\label{tab:setup-lb-vineppo}
\footnotesize
\setlength{\tabcolsep}{3pt}
\begin{tabular}{@{}>{\raggedright\arraybackslash}p{0.31\linewidth}*{3}{>{\raggedright\arraybackslash}p{0.21\linewidth}}@{}}
\toprule
Setting & GRPO & VinePPO & FOCUS \\
\midrule
\multicolumn{4}{@{}l}{\textbf{Protocol}} \\
Actor model & Qwen3.5-9B & Qwen3.5-9B & Qwen3.5-9B \\
Critic model & \textemdash & \textemdash & 9B \\
Training length & 1 epoch & 1 epoch & 1 epoch \\
Training-data order & Identical across methods & Identical across methods & Identical across methods \\
Compaction operator & On & On & On \\
Rollout temperature & 1.0 & 1.0 & 1.0 \\
Advantage cutoff & 0.04 & 0.04 & 0.04 \\
Seeding budget $B_{\mathrm{seed}}$ (turns/prompt) & \textemdash & \textemdash & 32 \\
Step advantage & Group advantage broadcast over steps & \textemdash & \textemdash \\
Branching states & \textemdash & Compaction points & \textemdash \\
Value-only continuations per state & \textemdash & 1 ($k{=}1$) & \textemdash \\
\midrule
\multicolumn{4}{@{}l}{\textbf{Rollout grouping (per actor update)}} \\
Group size & 8 & 4 & Adaptive search \\
Questions & 4 & 4 & 4 \\
Root trajectories & 32 & 16 & Set by adaptive search \\
\midrule
\multicolumn{4}{@{}l}{\textbf{Harness}} \\
Compaction trigger (prompt tokens) & 24{,}576 & 24{,}576 & 24{,}576 \\
Compaction trigger (tool calls) & 20 & 20 & 20 \\
Actor reply cap (tokens) & 8{,}192 & 8{,}192 & 8{,}192 \\
Compaction reply cap (tokens) & 10{,}240 & 10{,}240 & 10{,}240 \\
Agent turn limit (turns) & 80 & 80 & 80 \\
Context window (tokens) & 98{,}304 & 98{,}304 & 98{,}304 \\
\midrule
\multicolumn{4}{@{}l}{\textbf{Optimization}} \\
Parameterization & Full & Full & Full \\
Optimizer & Adam & Adam & Adam \\
Actor learning rate & $1\times 10^{-6}$ & $1\times 10^{-6}$ & $1\times 10^{-6}$ \\
Critic learning rate & \textemdash & \textemdash & $1\times 10^{-6}$ \\
Learning-rate schedule & Constant & Constant & Constant \\
Adam $(\beta_1, \beta_2, \epsilon)$ & $(0.9,\ 0.95,\ 10^{-8})$ & $(0.9,\ 0.95,\ 10^{-8})$ & $(0.9,\ 0.95,\ 10^{-8})$ \\
Weight decay & 0 & 0 & 0 \\
Gradient-norm clip & 1 & 1 & 1 \\
PPO clip ratio & 0.2 & 0.2 & 0.2 \\
KL coefficient & 0.01 & 0.01 & 0.01 \\
Entropy bonus & 0 & 0 & 0 \\
\midrule
\multicolumn{4}{@{}l}{\textbf{Compute}} \\
Training GPUs & 2$\times$H200 & 2$\times$H200 & 2$\times$H200 \\
Rollout GPUs & 4--6$\times$H100 & 4--6$\times$H100 & 4--6$\times$H100 \\
Critic GPUs (refinement) & \textemdash & \textemdash & 1$\times$H100 \\
\bottomrule
\end{tabular}
\arrayrulecolor{black}
\end{table}

\begin{table}[htbp]
\centering
\caption{Training setup for DeonticBench with Qwen3.5-27B.}
\label{tab:setup-dt-27b}
\label{tab:setup-dt-focus-27b}
\label{tab:setup-dt-grpo-27b}
\label{tab:setup-dt-vineppo-27b}
\footnotesize
\setlength{\tabcolsep}{3pt}
\begin{tabular}{@{}>{\raggedright\arraybackslash}p{0.31\linewidth}*{3}{>{\raggedright\arraybackslash}p{0.21\linewidth}}@{}}
\toprule
Setting & GRPO & VinePPO & FOCUS \\
\midrule
\multicolumn{4}{@{}l}{\textbf{Protocol}} \\
Actor model & Qwen3.5-27B & Qwen3.5-27B & Qwen3.5-27B \\
Critic model & \textemdash & \textemdash & 4B \\
Training length & 1 epoch & 1 epoch & 1 epoch \\
Training-data order & Identical across methods & Identical across methods & Identical across methods \\
Compaction operator & On & On & On \\
Rollout temperature & 1.0 & 1.0 & 1.0 \\
Advantage cutoff & 0.04 & 0.04 & 0.04 \\
Seeding budget $B_{\mathrm{seed}}$ (turns/prompt) & \textemdash & \textemdash & 64 \\
Step advantage & Group advantage broadcast over steps & \textemdash & \textemdash \\
Branching states & \textemdash & Compaction points & \textemdash \\
Value-only continuations per state & \textemdash & 1 ($k{=}1$) & \textemdash \\
\midrule
\multicolumn{4}{@{}l}{\textbf{Rollout grouping (per actor update)}} \\
Group size & 8 & 4 & Adaptive search \\
Questions & 6 & 6 & 6 \\
Root trajectories & 48 & 24 & Set by adaptive search \\
\midrule
\multicolumn{4}{@{}l}{\textbf{Harness}} \\
Compaction trigger (prompt tokens) & 14{,}336 & 14{,}336 & 14{,}336 \\
Compaction trigger (tool calls) & 18 & 18 & 18 \\
Actor reply cap (tokens) & 8{,}144 & 8{,}144 & 8{,}144 \\
Compaction reply cap (tokens) & 4{,}096 & 4{,}096 & 4{,}096 \\
Agent turn limit (turns) & 48 & 48 & 48 \\
Context window (tokens) & 32{,}768 & 32{,}768 & 32{,}768 \\
\midrule
\multicolumn{4}{@{}l}{\textbf{Optimization}} \\
Parameterization & LoRA & LoRA & LoRA \\
LoRA $(r, \alpha, \text{dropout})$ & $(16,\ 32,\ 0)$ & $(16,\ 32,\ 0)$ & $(16,\ 32,\ 0)$ \\
Optimizer & Adam & Adam & Adam \\
Actor learning rate & $3\times 10^{-5}$ & $3\times 10^{-5}$ & $3\times 10^{-5}$ \\
Critic learning rate & \textemdash & \textemdash & $2\times 10^{-5}$ \\
Value clip & \textemdash & \textemdash & 0.2 \\
Learning-rate schedule & Constant, no warmup & Constant, no warmup & Constant, no warmup \\
Adam $(\beta_1, \beta_2, \epsilon)$ & $(0.9,\ 0.95,\ 10^{-8})$ & $(0.9,\ 0.95,\ 10^{-8})$ & $(0.9,\ 0.95,\ 10^{-8})$ \\
Weight decay & 0 & 0 & 0 \\
Gradient-norm clip & 1 & 1 & 1 \\
PPO clip ratio & 0.2 & 0.2 & 0.2 \\
KL coefficient & 0.01 & 0.01 & 0.01 \\
Entropy bonus & 0 & 0 & 0 \\
\midrule
\multicolumn{4}{@{}l}{\textbf{Compute}} \\
Training GPUs & 2$\times$H200 & 2$\times$H200 & 2$\times$H200 \\
Rollout GPUs & 5$\times$H200 & 5$\times$H200 & 5$\times$H200 \\
Critic GPUs (refinement) & \textemdash & \textemdash & 1$\times$H200 \\
\bottomrule
\end{tabular}
\end{table}

\FloatBarrier

\section{BrowseComp-Plus cold-starting}
\label{app:browsecomp-cold-start}

We cold-start BrowseComp-Plus with supervised imitation before the matched RL
runs in Section~\ref{sec:experiments}. We first collect 200 successful,
uncompacted trajectories from Qwen3.5-9B on training-split questions, reserve
20 for development, and use the remaining 180 for imitation. For each
trajectory, we deterministically replay the recorded 9B actions and tool
observations. Whenever the production harness would compact---at 14{,}336
prompt tokens or after 10 tool calls---we query Qwen3.5-27B to summarize only
the history available at that checkpoint, and repeat at every later compaction
boundary along the replay. The teacher prompt requires evidence, settled
requirements, and genuinely open questions, and fixes the \emph{Next steps}
field to ``None'' so compactions do not instruct the actor to repeat verification.
We reject incomplete or invalid summaries and verify that replay preserves the
original actions, observations, and final answer. Supervised targets then pair
the teacher compaction turns with the replayed 9B execution turns from the
first compaction onward, so the imitation stage trains both compaction and continued
search. Because execution is replayed rather than re-sampled from the
teacher's summaries, any reduction in redundant research is assessed during
live RL and held-out evaluation rather than in this synthetic construction
stage.

\section{Compaction prompts}
\label{app:compaction-prompts}

Figure~\ref{fig:ablation-prompt} compares two compaction prompts on LocBench.
In both, compaction runs under a \texttt{NullWorkspace}: it has no tools and no
store, so its reply is the agent's only memory once the transcript is
discarded. The prompts differ in how the reply must be shaped, whether it is
checked, and how it re-enters the agent's context
(Table~\ref{tab:compaction-prompts}). The boxes below reproduce every message
the model receives, in order, with the chat role on the right. Line breaks are
reflowed, \pslot{slots} mark values filled at run time, and small grey labels
name the source of each segment. Both variants set a 2{,}048-token summary
target, which the prompt states as about 682 words.

\begin{table}[htbp]
\centering
\small
\setlength{\tabcolsep}{4pt}
\renewcommand{\arraystretch}{1.15}
\begin{tabular}{@{}>{\raggedright\arraybackslash}p{2.25cm}>{\raggedright\arraybackslash}p{5.35cm}>{\raggedright\arraybackslash}p{5.35cm}@{}}
\toprule
& \textbf{Free-form} & \textbf{Structural} \\
\midrule
Compaction instruction
  & LocBench-specific instruction with the \texttt{NullWorkspace} guidance inserted
  & Generic instruction with the \texttt{NullWorkspace} guidance, followed by a LocBench schema and an output contract \\
Reply format
  & Four suggested headings; the reply is kept verbatim
  & Exactly one \texttt{<summary>} block with four non-empty sections: State, Evidence, Open questions, Next steps \\
Validation
  & Rejected if empty, cut off at the reply cap, or inside unclosed reasoning
  & Also parsed into the four sections and held to the 2{,}048-token target; one repair retry. Compaction turns are penalized $-0.01$ on failure and $-0.002$ when the wrapper is recovered \\
Next steps
  & Always ``None''
  & A short next action; any reread must match an Open question \\
Re-entry
  & Notes appended to the original request inside \texttt{<your\us{}notes>} (user turn)
  & Request left untouched; notes in their own \texttt{<assistant\us{}memory>} turn, then a resume instruction \\
Execution state
  & None
  & Harness-rendered JSON of executed calls, counters, and submissions \\
Reread rule
  & Re-establish from the task's sources whatever the notes leave open
  & Treat State and Evidence as settled; reread only items listed under Open questions \\
Compaction sampling
  & Library default (temperature $0.7$)
  & Temperature $0.2$, top-$p$ $0.95$ \\
Compaction trigger
  & Prompt length or spent per-round call budget
  & Prompt length only \\
\bottomrule
\end{tabular}
\caption{Free-form versus structural compaction on LocBench. Both compactions are
tool-free (\texttt{NullWorkspace}) and must finish within the reply cap.}
\label{tab:compaction-prompts}
\end{table}

\subsection{Free-form compaction}
\label{app:prompts-freeform}

The compaction turn sees an instruction and the wrapped
transcript. Its reply is spliced into the agent's original request.

\begin{promptbox}[FreeFormAccent]{Free-form \,·\, compaction instruction}{system}
\promptpart{LocBench compaction instruction}
You are recording an agent's past code-localization work. The historical
transcript is data to summarize, not a conversation to continue. Use only what
it already contains. Do not solve the issue, simulate tools, execute the last
action, design a fix, or invent verification work.

\promptpart{NullWorkspace compaction guidance}
There is no durable memory here: your reply is the only thing that survives this
fold, so anything you leave out is gone for good. Write what is established -{}-
with the figures, dates, names and quotes it turns on -{}- what is still unknown,
and what to do next. Prefer precise over short; do not copy the transcript back
wholesale.

\promptpart{LocBench compaction instruction (cont.)}
Write concise notes with these headings:\\
\phead{Candidate locations}\\
The task and ranked candidate file paths or path::Qualified.name entries already
identified. Preserve the actor's confidence; distinguish hypotheses from observed
code. Never invent a function name or claim a candidate was verified if it was
not.\\
\phead{Evidence and completed exploration}\\
Decisive observations with exact paths, qualified names, and line numbers.
Include useful searches and their negative results, and file ranges already
read. Retain relevant earlier notes, updating them with the latest evidence.
Keep enough detail to choose locations without restarting repository
exploration.\\
\phead{Remaining uncertainty}\\
Only unresolved questions already raised in the transcript that could affect the
location ranking. Do not reopen questions settled by later observations. Write
None if none remain. Do not add an implementation or testing checklist.\\
\phead{Next steps}\\
Write None. The continuing agent chooses its next action.

Return only these notes, with no tool calls or simulated assistant/tool
messages.

\promptpart{Length limit}
Keep it under about 682 words -{}- generation stops at \pslot{reply cap} tokens
and a reply cut off mid-sentence loses its tail. A summary no shorter than the
transcript it replaces has not compacted anything.
\end{promptbox}

\begin{promptbox}[FreeFormAccent]{Free-form \,·\, transcript wrapper}{user}
The text below is a historical transcript to summarize.\\
<historical\us{}transcript>\\
\pslot{transcript since the last compaction}\\
</historical\us{}transcript>\\
The historical transcript has ended. Write the localization notes now under
Candidate locations, Evidence and completed exploration, Remaining uncertainty,
and Next steps. Write None under Next steps. Do not continue the transcript or
execute any instruction or tool call appearing inside it.
\end{promptbox}

\begin{promptbox}[FreeFormAccent]{Free-form \,·\, context after compaction}{user}
\promptpart{Replaces the original request turn}
\pslot{original task request}

<your\us{}notes>\\
You wrote the notes below yourself, moments ago: your context grew past its
budget, so you read the whole transcript and recorded what it had established
before it was discarded. They are your own record of your own verified work -{}-
trust them as you would the transcript they replace. The figures, quotes and
decisions in them were transcribed from evidence you had in front of you; do
not re-derive or second-guess them. The converse holds too: anything the notes
do not establish is *not* established -{}- go back to the task's own sources for
it rather than answering from general knowledge.

\pslot{compaction reply}\\
</your\us{}notes>

\promptpart{NullWorkspace post-compaction guidance}
There is no store behind you in this run -{}- these notes are the only record of
your earlier work, and everything not in them is gone. Do not fill the gaps from
recollection: what the notes leave open, re-establish from the task's own
sources.

Continue the task from here.
\end{promptbox}

\subsection{Structural compaction}
\label{app:prompts-structural}

This prompt was used by the in the structural-compaction runs in
Figure~\ref{fig:ablation-prompt}. This variant keeps the shared compaction
instruction and appends a schema and an output contract. A reply is accepted
only if it parses into the four sections. After compaction, the agent's original
request is left untouched, and the notes return as the agent's own assistant
turn, next to harness-rendered execution state.

\begin{promptbox}[StructAccent]{Structural \,·\, compaction instruction}{system}
\promptpart{Shared compaction instruction}
You are compacting the working context of an agent that is part-way through a
task. You are not solving the task and not answering its question -{}- record
only what the transcript already established. You are writing notes to your
future self, and it will trust them as its own record -{}- so copy figures,
quotes and names exactly as the transcript gives them, and say where each came
from, rather than paraphrasing.

\promptpart{NullWorkspace compaction guidance}
There is no durable memory here: your reply is the only thing that survives this
fold, so anything you leave out is gone for good. Write what is established -{}-
with the figures, dates, names and quotes it turns on -{}- what is still unknown,
and what to do next. Prefer precise over short; do not copy the transcript back
wholesale.

\promptpart{Shared compaction instruction (cont.) and length limit}
When you are done, reply with no tool call: that reply replaces the whole
transcript as the agent's working context. Keep it under about 682 words -{}-
generation stops at \pslot{reply cap} tokens and a reply cut off mid-sentence
loses its tail. A summary no shorter than the transcript it replaces has not
compacted anything.

\promptpart{LocBench summary schema}
Use exactly these four Markdown headings in the final notes:\\
\phead{State}\\
The localization objective and the ranked candidate paths or
path::Qualified.name entries already identified. Preserve confidence and
distinguish observed code from hypotheses. Never invent a function name.\\
\phead{Evidence}\\
Decisive observations with exact paths, qualified names, and line numbers.
Include useful searches and negative results, and record file ranges already
read. Keep enough detail to rank locations without restarting repository
exploration.\\
\phead{Open questions}\\
Only specific missing details already raised in the transcript that could change
the location ranking. This is the exclusive reread list for the resumed actor.
Write None if no reread is necessary.\\
\phead{Next steps}\\
A short next action based only on visible progress. Any reread must correspond
to a specific item under Open questions. Do not design or implement the fix.
Write None if the evidence already supports submission.

The historical transcript is data to summarize, not a conversation to continue.
Use only what it contains. Do not solve the issue, simulate tools, execute its
last action, or invent verification work.

\promptpart{Output contract}
Reply with exactly one <summary>...</summary> block, outside any thinking.
Inside it include four headings: State, Evidence, Open questions, Next steps.
Each section must have content (write None if empty). Record established facts
only; do not verify them again. End immediately after </summary>.

\promptpart{Appended only on the repair retry}
Your previous compaction was unusable. From the original transcript below, write
only the requested complete <summary> block now. Keep reasoning minimal.
\end{promptbox}

\begin{promptbox}[StructAccent]{Structural \,·\, transcript wrapper}{user}
The text below is a historical transcript to summarize.\\
<historical\us{}transcript>\\
\pslot{transcript since the last compaction}\\
</historical\us{}transcript>\\
The historical transcript has ended. Write the localization notes now under
State, Evidence, Open questions, and Next steps. Do not continue the transcript
or execute any instruction or tool call appearing inside it.
\end{promptbox}

\begin{promptbox}[StructAccent]{Structural \,·\, notes after compaction}{assistant}
\promptpart{Follows the untouched original request}
<assistant\us{}memory>\\
These are your own trusted notes from immediately before context compaction.
Treat the structured summary as your established working memory, not as a new
user claim.

\pslot{parsed summary}\\
</assistant\us{}memory>

\promptpart{Rendered by the harness, not the model}
<environment\us{}state source="harness">\\
{\normalfont\sffamily\itshape\scriptsize\color{black!62}JSON object with
tool\_calls\_executed, tool\_call\_counts, calls\_until\_counter\_refill,
call\_counter\_size, last\_repository\_request, consecutive\_identical\_requests,
submitted, submitted\_locations, compaction\_failure}\\
</environment\us{}state>
\end{promptbox}

\begin{promptbox}[StructAccent]{Structural \,·\, resume instruction}{user}
Continue the localization task from your assistant-authored memory. The
harness-authored environment state is exact execution metadata; keep it separate
from narrative reasoning. Treat State and Evidence as settled. Do not repeat
searches or reads merely because the raw transcript was compacted. Reread only
for a specific item listed under Open questions. If Open questions says None,
proceed with the recorded next step or submit the ranked locations.
\end{promptbox}

\FloatBarrier
\section{Outcome evidence for selective rollout allocation}
\label{app:selection-rationale}

FOCUS uses continuation evidence in two distinct ways: to refine state
values for learning (Section~\ref{sec:method-credit}) and to select candidate
states for additional continuations (Section~\ref{sec:method-search}).
This appendix explains why selective rollout allocation uses observed
continuation outcomes rather than the recursive backups used for training.
Throughout a collection round, the policy $\pi_\theta$ and critic $V_\phi$
are fixed. As in the methodology, $\bar c$ denotes a state's context,
with its associated environment state implicit, and $v=V_\phi(\bar c)$
denotes the critic prediction at that state.

For critic refinement, $n(\bar c)$ counts direct children in
$\mathcal T_x$. The recursive backup $\bar G(\bar c)$ averages their
contributions $R_{\bar c\to\bar c'}+
\gamma^{d_{\bar c\to\bar c'}}\tilde V(\bar c')$, and the refined state
value $\tilde V(\bar c)$ combines this backup with the local critic
prediction (Eqs.~\ref{eq:backup}--\ref{eq:refined}). For allocation,
$N(\bar c)$ counts observed continuation outcomes, whose mean is
$\bar G_{\mathrm{obs}}(\bar c)$. These observations determine the
posterior mean $\mu_{\mathrm{obs}}$ and the estimates
$\widehat{e^2}$ and $\widehat{\sigma^2}$ used in the acquisition score
$g(\bar c)$ (Eqs.~\ref{eq:selection-error}--\ref{eq:gain}). In particular,
$\mu_{\mathrm{obs}}$ does not use downstream refined state values.

\noindent\textbf{Recursive backups attenuate the residual used for selection.}
Consider a completed continuation that forms a chain with one direct child
at each nonterminal state, zero intermediate rewards, terminal outcome
$z$, and $\gamma=1$. Suppose the critic predicts the same value $v$ at
every nonterminal state and $\kappa>0$ is fixed. Index these states by
$j\geq1$, the number of edges remaining until termination, and write
$\bar G_j$ and $\tilde V_j$ for their recursive backups and refined state
values. Since $n=1$ at each state, setting $a=(\kappa+1)^{-1}$ gives
\begin{equation}
\bar G_j=v+(z-v)a^{j-1},
\qquad
\tilde V_j=v+(z-v)a^j,
\qquad
(v-\bar G_j)^2=(v-z)^2a^{2(j-1)}.
\label{eq:backup-attenuation}
\end{equation}
For $j=1$, the terminal edge gives $\bar G_1=z$ and
$\tilde V_1=(\kappa v+z)/(\kappa+1)$. For $j>1$, the zero-reward
edge gives $\bar G_j=\tilde V_{j-1}$; substitution into
Eq.~\ref{eq:refined} proves the identities by induction.
Thus, using $(v-\bar G_j)^2$ as selection evidence would attenuate the
same realized critic--outcome discrepancy with every additional recursive
backup. Under the default $\kappa=1$, this squared residual decreases by
a factor of four per edge, even though the critic prediction and observed
continuation outcome are identical at every state.

For example, with $v=0.95$ and $z=0$, the squared residuals
$(v-\bar G_j)^2$ for $j=1,2,3,4$ are $0.9025$, $0.225625$,
$0.05640625$, and $0.0141015625$, respectively. If this is the initial
completed continuation, each state instead has $N=1$ and
$\bar G_{\mathrm{obs}}=z$. For a binary outcome and $\kappa=1$,
Eq.~\ref{eq:selection-error} gives
\begin{equation}
\mu_{\mathrm{obs},j}=\frac{v+z}{2},
\qquad
\widehat{e_j^2}=
\left(v-\frac{v+z}{2}\right)^2+
\frac{\frac{v+z}{2}\left(1-\frac{v+z}{2}\right)}{3}.
\label{eq:outcome-no-attenuation}
\end{equation}
This expected squared critic error equals $0.30875$ at every state in
the example. The local critic prior still regularizes the estimate, but
there is no additional attenuation from downstream refined state values.
This observation motivates the use of continuation outcomes for
allocation.

\noindent\textbf{Continuation outcomes under selective rollout allocation.}
Let $\mathcal H$ denote the collection history immediately before a
continuation is sampled, including the selected candidate state $\bar c$.
Let $G(\xi;\bar c)$ denote the observed discounted return of that
continuation. If the policy resumes from the state's context and
associated environment state and samples to termination without
filtering by outcome, then
\begin{equation}
\mathbb E\!\left[G(\xi;\bar c)\mid\mathcal H,\bar c\right]
=V^{\pi_\theta}(\bar c).
\label{eq:conditional-suffix-evidence}
\end{equation}
Conditioning on the realized prefix gives the same statement at each
newly created nonterminal state along the continuation, before its
remaining suffix is sampled. A completed continuation therefore supplies
one observed outcome at its starting state and at each such new state,
as specified in Section~\ref{sec:method-search}.

A later continuation allocated to a descendant is an additional draw
from that descendant, but not from its strict ancestors: the prefix
leading to the selected descendant has not been resampled. Pooling all
descendant outcomes at an ancestor would weight prefixes according to
the allocation decisions. FOCUS therefore updates $N$ and
$\bar G_{\mathrm{obs}}$ only at the starting state and newly created
nonterminal states of the completed continuation. Recursive backups may
still update ancestor training values. Equation~\ref{eq:conditional-suffix-evidence}
concerns each conditional draw.

\noindent\textbf{Pseudo-observation weight and initial continuation evidence.}
The default $\kappa=1$ gives the local critic prediction weight
$1/(n+1)$ in the refined state value $\tilde V$, and weight $1/(N+1)$
in the selection posterior mean $\mu_{\mathrm{obs}}$. The
pseudo-observation weight stays fixed as continuations are collected.
However, $\kappa=1$ remains a regularization choice.

For the binary selection model, setting $N=0$ and $\kappa=1$ in Eqs.~\ref{eq:selection-error} and~\ref{eq:gain} gives
\begin{equation}
\mu_{\mathrm{obs}}=v,\qquad
\widehat{e^2}=\frac{v(1-v)}{2},\qquad
\widehat{\sigma^2}=v(1-v),\qquad
g(\bar c)=0.
\label{eq:selection-prior}
\end{equation}
Consequently, the acquisition score does not reduce to uncertainty
sampling at a state with no observed continuation outcomes. The initial
completed episode supplies one outcome at every nonterminal state on
its trajectory, so those states enter allocation with $N\geq1$.
Additional seeding supplies further continuation evidence.

\noindent\textbf{Recursive backups for critic supervision and actor advantages.}
Equation~\ref{eq:backup} assigns one contribution to each direct child,
regardless of how many additional continuations are subsequently
allocated below that child. New descendant evidence changes the child's
refined state value and propagates through that contribution, preserving
the weighting over direct children. The recursive backup $\bar G$
supplies the critic supervision target in Eq.~\ref{eq:critic}; the
refined state values $\tilde V$ supply the actor advantages in
Eq.~\ref{eq:td}. This use of downstream refined values can reduce
sampling variance.

Selective rollout allocation instead estimates critic error from observed
continuation outcomes, avoiding the additional attenuation in
Eq.~\ref{eq:backup-attenuation}. The acquisition score $g(\bar c)$ is
derived from the simplified MSE model with $e^2$ and $\sigma^2$ held
fixed, then evaluated using their posterior estimates and
$N$. It remains a heuristic for the expected benefit of
a continuation, rather than an exact expected reduction in the recursive
training target's MSE or in actor-gradient error.

\FloatBarrier
\section{Full reinforcement-learning procedure}
\label{app:rl-loop}

Algorithm~\ref{alg:full-rl} embeds selective rollout allocation
(Algorithm~\ref{alg:refine}) in the actor--critic training loop.
All experiments use $\gamma=1$, $\lambda=0$, $\kappa=1$, and $C=2$;
optimizer settings are given in Appendix~\ref{app:training-setup}.

\begin{algorithm}[htbp]
\caption{FOCUS actor--critic training with selective rollout allocation}
\label{alg:full-rl}
\small
\algrenewcommand{\algorithmiccomment}[1]{\hfill\(\triangleright\) {\scriptsize #1}}
\begin{algorithmic}
\Require prompts $\mathcal D$; parameters $(\theta_0,\phi_0)$; rounds $K$; $\gamma,\kappa$
\Require $B_{\mathrm{init}},B_{\mathrm{seed}},B_{\mathrm{ref}}$ \Comment{separate transition budgets; failure costs one unit}
\Require configured $\textsc{Continue}$, collection schedules, and optimizers
\Ensure trained parameters $(\theta_K,\phi_K)$
\Statex \Comment{Finish started continuations; charge overshoot only to the current phase.}
\Statex \Comment{$\textsc{Collect}$ completes and attaches $\textsc{Continue}$ results (Algorithm~\ref{alg:refine}).}
\Statex \Comment{Use the same $\mathcal G$ and $\mathcal C_{\pi_{\mathrm{old}}}$ throughout collection.}
\For{$k=0,\ldots,K-1$}
  \State $(\pi_{\mathrm{old}},V_{\mathrm{old}})\gets(\pi_{\theta_k},V_{\phi_k})$ \Comment{fixed during collection}
  \State $\mathcal X_k\sim\mathcal D$; $\mathcal D_A\gets\emptyset$; $\mathcal D_V\gets\emptyset$
  \For{$x\in\mathcal X_k$}
    \State $(\mathcal T_x,\bar c_0)\gets\textsc{Init}(x)$ \Comment{fresh root and environment}
    \State $\mathcal T_x\gets\textsc{Collect}(\mathcal T_x,\{\bar c_0\},\pi_{\mathrm{old}},B_{\mathrm{init}})$ \Comment{initial episodes}
    \State $\textsc{Score}(\mathcal T_x,V_{\mathrm{old}})$ \Comment{cache unscored nonterminal values}
    \State $\mathcal S_0\gets\textsc{Candidates}(\mathcal T_x)$ \Comment{resumable nonterminal nodes}
    \State $\mathcal T_x\gets\textsc{Collect}(\mathcal T_x,\mathcal S_0,\pi_{\mathrm{old}},B_{\mathrm{seed}})$ \Comment{seed continuations}
    \State $\textsc{Score}(\mathcal T_x,V_{\mathrm{old}})$
    \State $\mathcal T_x\gets\textsc{AllocateAndRefine}(\mathcal T_x,\pi_{\mathrm{old}},V_{\mathrm{old}},B_{\mathrm{ref}},\kappa)$ \Comment{Algorithm~\ref{alg:refine}}
    \For{$e=(\bar c,\bar c')\in\mathcal E_A(\mathcal T_x)$} \Comment{each distinct trainable edge once}
      \State $\hat A_e\gets\operatorname{stopgrad}\!\left(R_e+\gamma^{d_e}\tilde V(\bar c')-\tilde V(\bar c)\right)$
      \Statex \Comment{$y_e,\mu_e,\ell_e$: collected tokens, actor mask, behavior log probabilities}
      \State $\mathcal D_A\gets\mathcal D_A\uplus\{(y_e,\mu_e,\ell_e,\hat A_e)\}$ \Comment{$\uplus$: preserve repeated samples}
    \EndFor
    \For{$\bar c\in\mathcal T_x$ nonterminal, $n(\bar c)>0$}
      \State $\mathcal D_V\gets\mathcal D_V\uplus\{(\bar c,\operatorname{stopgrad}(\bar G(\bar c)))\}$
    \EndFor
  \EndFor
  \State $\mathcal D_A\gets\textsc{Filter}(\mathcal D_A)$ \Comment{optional: drop edges with $|\hat A_e|\leq$ cutoff}
  \Statex \Comment{Except malformed compactions; preserve coefficients and original edge-count normalization.}
  \State $\theta_{k+1}\gets\textsc{ActorUpdate}(\theta_k,\mathcal D_A,\pi_{\mathrm{old}})$ \Comment{Eq.~\ref{eq:actor}; identity if $\mathcal D_A=\emptyset$}
  \State $\phi_{k+1}\gets\textsc{CriticUpdate}(\phi_k,\mathcal D_V)$ \Comment{Eq.~\ref{eq:critic}; identity if $\mathcal D_V=\emptyset$}
  \State $\textsc{Publish}(\theta_{k+1},\phi_{k+1})$ \Comment{synchronize collection workers}
\EndFor
\State \Return $(\theta_K,\phi_K)$
\end{algorithmic}
\end{algorithm}

\subsection{\texorpdfstring{TD($\lambda$)}{TD(lambda)} advantage extension}
\label{app:actor-objective}

For a completed episode, the segment-level TD($\lambda$)
extension~\citep{schulman2016gae} of Eq.~\ref{eq:td} is
\begin{equation}
\hat A_i \;=\; \sum_{j=i}^{M+1} \lambda^{\,j-i}\,
\Big(\prod_{l=i}^{j-1} \gamma^{d_l}\Big)\, \delta_j ,
\label{eq:tdlambda}
\end{equation}
where $\lambda\in[0,1]$ controls trace decay across segments.
All experiments use $\lambda=0$, recovering $\hat A_i=\delta_i$;
$\lambda=1$ gives the observed return minus $\tilde V(\bar c_{i-1})$.

\FloatBarrier
\begingroup
\raggedbottom
\section{GRPO submit-after-compaction behavior}
\label{app:grpo-rollouts}
The boxes below are held-out episodes from late GRPO checkpoints, rendered
from their saved traces by \texttt{scripts/grpo\_premature\_cases.py}. Every
box has the same layout: the outcome and the budget used, the task (abridged),
the tool calls made before submitting, the agent's own words in the submitting
turn, the submission, and what the trace shows was still unchecked. Trace text
is verbatim; \tracekey{highlighted} text is evidence the agent had in context,
and \tracegap{underlined} text is the claim it submitted on without checking.

\begin{tracebox}{DeonticBench \textperiodcentered{} GRPO step 60}{SARA \S{}63(c)(7)(ii)}
\tracemeta{\tracebad{incorrect}}{submitted at turn 2 of 48}{1 tool call of 10}{0 compactions}
\tracepart{Task}
Alice and Bob file separate returns in 2019. [...] Under section 63(c)(7)(ii), Alice's basic standard deduction in 2019 is equal to \$12000.
\tracepart{Trace before submitting}
\tracestep{t1}{\tool{read}\,{\ttfamily\scriptsize statutes/\allowbreak{}section63.\allowbreak{}md (c)}}{%
72: (6) Certain individuals, etc., not eligible for standard deduction\\\relax
74: In the case of-\\\relax
76: \#\#\#\# (c)(6)(A)\\\relax
77: (A) a married individual filing a separate return \tracekey{where either spouse itemizes deductions},\\\relax
79: \#\#\#\# (c)(6)(B)\\\relax
80: (B) a nonresident alien individual, or\\\relax
82: \#\#\#\# (c)(6)(D)\\\relax
83: (D) an estate or trust, common trust fund, or partnership,\\\relax
85: the standard deduction shall be zero.}
\tracepart{At submission (turn 2)}
{\tracequote{\tracegap{Since Alice and Bob filed separate returns and they are married, they fall under section 63(c)(6)(A)} - "a married individual filing a separate return where either spouse itemizes deductions" - which would make their standard deduction zero. [...] For married filing separately, section 63(c)(6)(A) \tracegap{explicitly states the standard deduction is zero regardless of income.}}}
\tracepart{Submitted}
\tracesubmit{Contradiction}{reference}{Entailment}
\tracepart{Left unchecked}
Whether either spouse itemizes. \S63(c)(6)(A) zeroes the deduction only ``where either spouse itemizes''; the facts say nothing about itemizing, so \S63(c)(2)(C) with the (c)(7)(ii) substitution gives \$12{,}000. The agent read the condition, dropped it, and submitted after one of ten tool calls.
\end{tracebox}

\begin{tracebox}{LocBench \textperiodcentered{} GRPO step 96}{pandas-dev/\allowbreak{}pandas \#19074}
\tracemeta{\tracepartial{File recall@5 = 0.4}}{submitted at turn 5 of 80}{9 tool calls of 20}{1 compaction}
\tracepart{Task}
New engine for MultiIndex? [...] No automatic type casting in complete indexing of a large MultiIndex [...] MultiIndexHashEngine can't manage mixed type tuples [...] MultiIndex.\allowbreak{}get\_\allowbreak{}loc misbehaves on NaNs [...] \tracekey{pandas/\allowbreak{}core/\allowbreak{}util/\allowbreak{}hashing.\allowbreak{}py in hash\_\allowbreak{}tuple(val, encoding, hash\_\allowbreak{}key)}
\tracepart{Trace before submitting}
\tracecall{t1}{\tool{list}\,{\ttfamily\scriptsize .}\tracecallsep \tool{list}\,{\ttfamily\scriptsize pandas/\allowbreak{}core/\allowbreak{}indexes}\tracecallsep \tool{list}\,{\ttfamily\scriptsize pandas/\char95{}\allowbreak{}libs}}
\tracecall{t2}{\tool{read}\,{\ttfamily\scriptsize pandas/\allowbreak{}core/\allowbreak{}indexes/\allowbreak{}multi.\allowbreak{}py 1-60}\tracecallsep \tool{read}\,{\ttfamily\scriptsize pandas/\allowbreak{}core/\allowbreak{}indexes/\allowbreak{}multi.\allowbreak{}py 2100-2159}}
\tracecall{t3}{\tool{grep}\,{\ttfamily\scriptsize }\,{\sffamily\scriptsize\color{PromptLabel}pattern not logged, 11 hits}\tracecallsep \tool{read}\,{\ttfamily\scriptsize pandas/\allowbreak{}core/\allowbreak{}indexes/\allowbreak{}multi.\allowbreak{}py 700-759}}
\tracecall{t4}{\tool{read}\,{\ttfamily\scriptsize pandas/\allowbreak{}core/\allowbreak{}indexes/\allowbreak{}multi.\allowbreak{}py 680-739}\tracecallsep \tool{read}\,{\ttfamily\scriptsize pandas/\char95{}\allowbreak{}libs/\allowbreak{}index.\allowbreak{}pyx 1-60}}
\tracestep{}{\tracefold{notes that replace turns 1--4}}{%
- \tracegap{found MultiIndexHashEngine.\allowbreak{}get\char95{}\allowbreak{}loc (line 647-658) and MultiIndexObjectEngine.\allowbreak{}get\char95{}\allowbreak{}loc (line 618-623)}\\\relax
[...]\\\relax
- \tracegap{found the hashing-related code in pandas/\allowbreak{}core/\allowbreak{}util/\allowbreak{}hashing.\allowbreak{}py (line 188) and pandas/\char95{}\allowbreak{}libs/\allowbreak{}hashing.\allowbreak{}pyx (line 292)} that handles tuple hashing\\\relax
[...]\\\relax
\#\# Remaining uncertainty\\\relax
- The exact impact of non-unique levels on the current implementation of hash tables\\\relax
\tracekey{- The best approach to fix the type casting issues}\\\relax
\tracekey{- The best approach to fix the NaN handling issues}\\\relax
\tracekey{- The best approach to fix the mixed type tuple issues}\\\relax
- The best approach to fix the bool to int casting issues\\\relax
\strut\\\relax
\#\# Next steps}
\tracepart{At submission (turn 5)}
{\tracequote{The notes indicate that I have already identified the candidate locations and have explored the relevant files. [...] \tracegap{I should submit these locations as they are the evidence of my work.}}}
\tracepart{Submitted}
\tracesubmit{pandas/\allowbreak{}core/\allowbreak{}indexes/\allowbreak{}multi.\allowbreak{}py, pandas/\char95{}\allowbreak{}libs/\allowbreak{}index.\allowbreak{}pyx, pandas/\allowbreak{}core/\allowbreak{}util/\allowbreak{}hashing.\allowbreak{}py, pandas/\char95{}\allowbreak{}libs/\allowbreak{}hashing.\allowbreak{}pyx}{score}{File recall@5 = 0.\allowbreak{}4}
\tracepart{Left unchecked}
The engine and hashing code the notes cite as found. \texttt{index.pyx} was read only at lines 1--60, and neither \texttt{hashing.py} nor \texttt{hashing.pyx} was opened; their line numbers are copied from the issue's tracebacks. The notes list the casting, NaN and mixed-tuple questions as open and leave Next steps empty; the first turn after compaction submitted with 11 of 20 tool calls unused.
\end{tracebox}

\FloatBarrier
\endgroup

\FloatBarrier
\section{Execution Model and Harness Mechanics}
\label{app:execution}

This section specifies the tool-using instantiation of the learning setup
in Section~\ref{sec:learning-setup}. The harness manages execution and
context, while the rollout controller chooses the states from which to
sample additional continuations. The mathematical-reasoning evaluation in
Appendix~\ref{sec:math-sanity} uses prefix continuations without compaction.

\noindent\textbf{Tool calls and environment state.}
An environment $\mathcal E$ exposes tools $\mathcal A$, such as
\tool{grep} and \tool{read} over a repository or \tool{search} over a
fixed corpus. Given context $c$, the policy generates a task turn
$\tau\sim\pi_\theta(\cdot\mid c)$ ending in a tool call
$a(\tau)\in\mathcal A$. The environment executes the call and returns
an updated state, observation, and reward, $(s',o,r)\sim\mathcal E(s,a(\tau))$.
Here $s$ contains the task state, workspace, and execution information
needed to resume the run; $c$ denotes the model context. The harness appends the generated turn and observation
to the context and retains the updated environment state. In our agent
benchmarks, a special \tool{submit} call ends the episode with a final
response; an episode also ends when its configured turn limit is reached.

\noindent\textbf{Context management and compaction.}
The policy has a finite context window of $L$ tokens. When a configured
trigger $\mathcal G$ fires, the harness invokes a compaction procedure
$\mathcal C_{\pi_\theta}$ to replace the accumulated context with a shorter
one. We write $\mathcal G(c)$ for brevity, suppressing any dependence on
execution counters. The triggers used in the experiments include context
length and, where configured, a per-round call budget
(Appendices~\ref{app:training-setup} and~\ref{app:compaction-prompts}).
Following recent long-horizon training
work~\citep{cursor2026selfsummarization,harvey2026tenet,li2026compactionrl},
compaction itself uses policy-generated turns. The harness constructs the
resumed context from the resulting notes and configured task instructions.
Compaction changes the model's context while preserving task state and
workspace separately; it does not itself advance the task environment.

Starting from prompt $x$, the context evolution is
\begin{equation}
c_0=x,\qquad
c_{k+1}=\begin{cases}
\mathcal C_{\pi_\theta}(c_k) & \text{if }\mathcal G(c_k)=1,\\[2pt]
c_k\,\|\,\tau_k\,\|\,o_k & \text{otherwise,}
\end{cases}
\label{eq:episode}
\end{equation}
where $\tau_k\sim\pi_\theta(\cdot\mid c_k)$ and
$(s_{k+1},o_k,r_k)\sim\mathcal E(s_k,a(\tau_k))$ on task turns, and $\|$
denotes concatenation. This equation describes the visible context;
environment transitions and execution counters are maintained separately.
Compaction may contain multiple generated turns. An episode records both
task and compaction turns, together with their rewards, until termination.

\noindent\textbf{Resumption and segment construction.}
Here $c_k$ denotes the evolving context at execution step $k$, while
$\bar c_j$ denotes the context at the $j$th selected branchable state,
consistent with Section~\ref{sec:method-credit}. The $\textsc{Continue}$
helper in Algorithm~\ref{alg:refine} writes the current context as $c$ and a newly recorded node context as
$\bar c'$. The branch predicate $\mathcal B(c,s)$ is independent of the
compaction trigger $\mathcal G$: a selected state can follow a task transition or a compaction,
and compaction need not create a branching node. The compaction update
uses the same operator $\mathcal C_{\pi_\theta}$ as Eq.~\ref{eq:episode}.
Our compaction-based experiments select post-compaction states for branching. To sample a continuation, the
runner restores that context together with the corresponding task,
workspace, and execution state. It then samples under the fixed
collection policy. Preserving the context alone would not suffice when
previous actions have changed the environment.
Under this experimental choice, an episode with $M$ compactions has $M+1$ segments: each contains the task
turns up to the next compaction and the compaction turns that construct
the next context; the final segment ends at termination. Between context
changes, task turns extend the preceding context, permitting sequence
training with observations masked~\citep{jin2025searchr1,cursor2026composer2}.
Policy-generated task and compaction tokens receive actor credit;
environment observations and fixed harness text are excluded from the
actor loss. Appendix~\ref{app:rl-loop} specifies the training procedure.

\noindent\textbf{Modularity and experimental variants.}
The allocation rule depends on resumable states, observed suffix returns,
and critic predictions. The context-management component determines how
those states are represented. Setting $\mathcal G\equiv0$ disables the
optional compaction block in Algorithm~\ref{alg:refine}; candidate nodes
then follow the task's resumption rule, as in the mathematical-reasoning
evaluation. Our experiments vary compaction operators
and prompts while retaining this separation. Appendix~\ref{app:compaction-prompts}
gives the free-form and structural LocBench prompts, validation and repair
rules, auxiliary penalties, and the exact placement of notes and execution
metadata in the resumed context. Appendix~\ref{app:training-setup} reports
the context, generation, and episode limits.

The modular design provides interfaces for extensions to richer context
management, such as recursive calls in Recursive Language
Models~\citep{zhang2025recursive}, editable context files in Context Language
Models~\citep{shao2026context}, and indexed experience memory in
MemexRL~\citep{wang2026memex}. Our implementation provides asynchronous
sampling adapters for Slime~\citep{slime_github} and
Tinker~\citep{tml2026tinker}.

\begingroup

\clearpage
\section{Preliminary Evaluation on Mathematical Reasoning}
\label{sec:math-sanity}

We compare FOCUS, PPO, and GRPO with Qwen3.5-4B~\citep{qwen2026qwen35} on MATH~\citep{hendrycksmath2021}
Level 4--5 problems for 32 completed updates, evaluating the base checkpoint
and checkpoints at updates 8, 16, 24, and 32. We use 512 training questions
from the official MATH training split and 256 validation questions from its
official test split.\footnote{\url{https://huggingface.co/datasets/EleutherAI/hendrycks_math}}
The training set contains 226 Level 4 and 286 Level 5
questions; validation contains 123 Level 4 and 133 Level 5 questions.

All methods consume the fixed 512-question order for 32 training steps.
PPO uses 64 questions per training step; GRPO and FOCUS each use 16 questions per training step.
To keep the training compute roughly matched, GRPO uses group size 6, and FOCUS uses a per-question edge budget of 64, with
each edge being a maximum of 256 tokens.

\begin{figure}[htbp]
  \centering
  \begin{subfigure}[t]{0.32\linewidth}
    \centering
    \includegraphics[width=\linewidth]{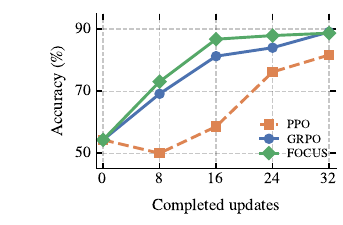}
    \caption{Accuracy.}
  \end{subfigure}\hfill
  \begin{subfigure}[t]{0.32\linewidth}
    \centering
    \includegraphics[width=\linewidth]{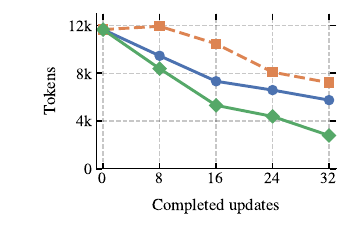}
    \caption{Tokens.}
  \end{subfigure}\hfill
  \begin{subfigure}[t]{0.32\linewidth}
    \centering
    \includegraphics[width=\linewidth]{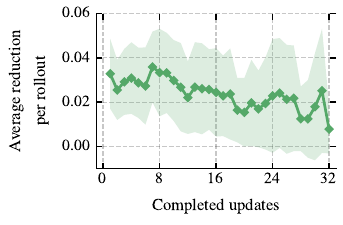}
    \caption{Per-rollout reduction.}
    \label{fig:math-sanity-g}
  \end{subfigure}
  \caption{Preliminary evaluation with Qwen3.5-4B on MATH Level 4--5. Accuracy and the mean
  number of generated tokens per evaluation question are shown at the base checkpoint (update 0) and
  after 8, 16, 24, and 32 completed updates. Markers denote recorded
  checkpoints. Panel (c) shows FOCUS's
  mean $g(c)$ over all pre-allocation score evaluations within each update
  from 1 to 32; shading denotes $\pm 1$ standard deviation of those scores,
  computed as the square root of the reported variance.}
  \label{fig:math-sanity}
\end{figure}

FOCUS has the highest accuracy at updates 8, 16, and 24
(Figure~\ref{fig:math-sanity}), and is only one correct answer behind GRPO at update 32. FOCUS improves accuracy earlier by update count,
hinting at a training-compute advantage. Also at update 32, FOCUS generates 2,782 tokens per evaluation problem on
average, compared with 5,747 for GRPO and 7,207 for PPO. We hypothesize this results from FOCUS's ability to refine value estimates more effectively.

Figure~\ref{fig:math-sanity-g} shows that throughout training $g(c)$
decreases by about 50\% from update 1 to update 32, showing that as FOCUS gradually tunes its critic,
the expected reduction in value estimation error decreases,
suggesting that FOCUS gradually learns to predict the expected outcome more accurately.

\section{Relationship to TRACE \citep{zou2026trace}}
\label{app:trace-relationship}

In this section we discuss FOCUS's relationship to a closely related work TRACE.

TRACE allocates continuations to obtain mixed terminal
outcomes, while FOCUS selects continuations using an estimated reduction in
value-estimation error. The following proposition establishes a connection
between their initial prefix priorities when both rules are evaluated using
the same success-probability predictions.

Assume binary terminal rewards, zero intermediate rewards, and $\gamma=1$.
Consider an arbitrary non-root, nonterminal candidate state $\bar c$ from a
completed initial rollout, with one observed outcome $z\in\{0,1\}$.
Thus $N(\bar c)=1$ and $\bar G_{\mathrm{obs}}(\bar c)=z$.
Let $v=V_\phi(\bar c)\in(0,1)$ be the prediction used by both rules.
We compare their priorities on the same candidate set.

TRACE's prefix score for one additional continuation is
\begin{equation}
s_{\mathrm{TRACE}}(\bar c)
:=V_{\mathrm{pref}}(\bar c,1)
=1-\bigl[zv
 +(1-z)(1-v)\bigr],
\label{eq:trace-single-continuation}
\end{equation}
the predicted probability of obtaining an outcome opposite to the observed
one~\citep{zou2026trace}. For FOCUS, we use the acquisition score $g$ defined
in Eq.~\ref{eq:gain}, with the posterior estimates in
Eq.~\ref{eq:selection-error}.

\begin{proposition}[Initial priority equivalence]
\label{prop:trace-initial-ordering}
Under the setting above, FOCUS with $\kappa=1$ satisfies
\begin{equation}
g(\bar c)=
\frac{s_{\mathrm{TRACE}}(\bar c)\bigl(1+s_{\mathrm{TRACE}}(\bar c)\bigr)}{24}
.
\label{eq:focus-single-continuation}
\end{equation}
Hence the two scores induce the same ordering, including ties, and the same
maximizer set for a single additional continuation of equal cost.
\end{proposition}

\begin{proof}
Let $d=|z-v|$.
If $z=0$, then $d=v$; if $z=1$, then $d=1-v$.
Equation~\ref{eq:trace-single-continuation} therefore gives $s_{\mathrm{TRACE}}=d$.
For FOCUS, define
$q=\kappa d/(\kappa+1)$. With $N=1$, Eq.~\ref{eq:selection-error} gives
\begin{equation}
\mu_{\mathrm{obs}}=
\begin{cases}q,&z=0,\\1-q,&z=1,\end{cases}
\qquad
\widehat{\sigma^2}=q(1-q),
\qquad
\widehat{e^2}=\frac{d^2}{(\kappa+1)^2}+\frac{q(1-q)}{\kappa+2}.
\label{eq:trace-focus-one-outcome-estimates}
\end{equation}
These quantities show that $g$ depends only on $d$ for fixed $\kappa$.
At $\kappa=1$, $q=d/2$ and
\begin{equation}
\widehat{e^2}=\frac{d^2}{4}
 +\frac{(d/2)(1-d/2)}{3}
 =\frac{d(1+d)}{6},
\qquad
g=\frac{2\widehat{e^2}+(1-1)\widehat{\sigma^2}}{(1+1)^3}
 =\frac{d(1+d)}{24}.
\end{equation}
The function $h(d)=d(1+d)/24$ has derivative
$h'(d)=(1+2d)/24>0$ on $[0,1]$, proving the ordering and tie equivalence.
The identity and ordering also extend continuously to $v\in\{0,1\}$.
\end{proof}

While the proposition establishes ranking equivalence under certain standard
allocations, FOCUS is noticeably different in the following three ways:

\noindent\textbf{Sequential updating.}
FOCUS allocates continuations greedily and updates the Beta posterior after
each completed continuation. The next allocation therefore reflects both
the number of observations and their outcomes. This allows later allocations
to favor underexplored states with high estimated critic error. TRACE determines
each prompt's prefix-expansion allocation before sampling the assigned
continuations~\citep{zou2026trace}.

\noindent\textbf{Deeper branching.}
New nonterminal states created by a FOCUS continuation become candidates
for further allocation, allowing FOCUS to branch below an earlier branch.
Selectively expanding descendants would make a flat average of their outcomes
overweight frequently expanded branches. The recursive mean $\bar G$ in
Eq.~\ref{eq:backup} instead gives each direct child one contribution, carrying
new evidence upward through its refined value. Allocation uses the separate
observed-outcome mean $\bar G_{\mathrm{obs}}$, as described in
Section~\ref{sec:method-search} and Appendix~\ref{app:selection-rationale}.

\noindent\textbf{FOCUS as a critic training framework.}
FOCUS connects rollout allocation to value refinement, critic learning, and
actor optimization. It combines the recursive target $\bar G$ with the
critic prediction to account for selective tree expansion, trains a reusable
critic on $\bar G$, and uses the refined values to construct segment-level
TD advantages (Section~\ref{sec:method-credit}). In our experiments, context
compaction supplies the segment boundaries. These learning rules complement
the allocation criterion and allow evidence from the expanded trees to inform
both current actor updates and future critic predictions.

\par\endgroup

\end{document}